\documentclass{article}

 \usepackage[preprint]{neurips_2026}

\usepackage[utf8]{inputenc} % allow utf-8 input
\usepackage[T1]{fontenc}    % use 8-bit T1 fonts
\usepackage{url}            % simple URL typesetting
\usepackage{booktabs}       % professional-quality tables
\usepackage{amsfonts}       % blackboard math symbols
\usepackage{nicefrac}       % compact symbols for 1/2, etc.
\usepackage{microtype}      % microtypography
\usepackage{xcolor}         % colors

\usepackage{amsmath}
\usepackage{pdfpages}
\usepackage{subcaption}
\usepackage{epstopdf}
\usepackage{epsfig}
\usepackage{graphicx}
\usepackage{multirow}
\usepackage{comment}
\usepackage{wrapfig}

\usepackage{algorithm}
\usepackage{algorithmic}

\usepackage{tabularx}

\usepackage{pgfgantt}
\usepackage{lscape}
\usepackage{ctable, makecell}
\usepackage{wrapfig, layouts}
\usepackage{inputenc}
\usepackage{geometry}
\usepackage{soul}

\newcommand{\ie}{\textit{i.e.}}
\newcommand{\eg}{\textit{e.g.}}

\definecolor{cvprblue}{rgb}{0.21,0.49,0.74}
\usepackage[pagebackref,breaklinks,colorlinks,citecolor=cvprblue]{hyperref}

\newcommand{\MYhref}[3][blue]{\href{#2}{\color{#1}{#3}}}

\usepackage{color, colortbl}
\title{AnomalyAgent: Training-Free Agentic Models for Zero-/Few-Shot Anomaly Detection}

\author{Yi Zhang*, Jiawen Zhu*, Lele Fu, Guansong Pang \\
Singapore Management University, Singapore \\
Sun Yat-sen University, China \\
\texttt{dylan.zhang9420@gmail.com}
}

\begin{document}

\maketitle

\begin{abstract}
Benefiting from generalizability of vision-language models (VLMs) such as CLIP, many zero-/few-shot anomaly detection (AD) approaches have achieved impressive detection performance across various datasets. Nevertheless, they require substantial training on large auxiliary datasets to adapt VLMs to anomaly detection, and their inference largely relies on visual–text embedding similarity-based anomaly scores, lacking reasoning abilities to detect complex anomalies that require in-depth contextual understanding. 
% their limited reasoning capability often restricts generalization to narrow domains and anomaly patterns (\eg, textual anomalies in industrial defects or medical lesion images), making them ineffective for anomalies that require deeper contextual understanding.
% practical value in both industrial and medical scenarios. While most existing approaches rely heavily on CLIP-based representations, 
% their generalization ability is often limited to narrow dataset types and predefined anomaly patterns, \eg, textual anomalies in industrial defect and medical legion images, failing to detect anomalies whose detection requires in-depth understanding of their context. 
% In contrast, the potential of multimodal large language models (MLLMs) for image anomaly detection remains largely unexplored. 
To address this limitation, we propose \textbf{AnomalyAgent}, a novel training-free, agentic framework that leverages the advanced reasoning and generalization capabilities of multimodal large language models (MLLMs) for anomaly detection. The key ingredients include \textbf{1)} a comprehensive anomaly-centric toolset that enables adaptive MLLM-driven, agentic anomaly reasoning in zero-shot settings, and \textbf{2)} a customized memory module that grounds anomaly reasoning with few-shot, in-context reference examples.
% Specifically, a ToolAgent is designed for zero-shot anomaly detection with a comprehensive toolset that integrates generic image processing operators with anomaly-aware analytical tools, enabling adaptive image enhancement and structured anomaly reasoning.
% Further, we introduce ToolAgent with structured vision memory banks that effectively collect and retrieve memories for few-shot anomaly detection, deriving the MemoryAgent. 
We extend evaluation beyond the detection of simple anomalies (\eg, surface defects like cracks and dents and clear legions) in widely used benchmarks to more diverse types of anomalies such as logical/contextual anomalies in logistics and manufacturing settings.
Extensive experiment results demonstrate that our AnomalyAgent achieves substantially better performance compared to training-free VLM-based AD and generic agentic methods, highlighting its superior generalization capability in both zero-shot and few-shot anomaly detection settings. The code implementation can be find at \MYhref{https://github.com/mala-lab/AnomalyAgent}{this address}.
\end{abstract}

\section{Introduction}
% P1: What are ZSAD and FSAD, and their potentials --> call for generalization and practicability
Anomaly detection (AD) is one key task in computer vision, aiming to identify instances that deviate significantly from the underlying data distribution. It is critical for a wide range of real-world applications, including industrial inspection, medical diagnosis, security monitoring, and autonomous systems~\cite{bergmann2019MVTec,salehi2021multiresolution,bergmann2022MVTecloco,hofer2025kaputt}.
A key challenge in AD lies in its inherent open-ended nature, \textit{ie}, anomalies are rare, diverse and often unpredictable, making it impractical to exhaustively model all possible abnormalites during training. To tackle this challenge, conventional approaches~\cite{pang2021Anomaly,cao2024survey,wu2026deep} focus on learning normal patterns from large normal training data instead. Recent advances have shifted from  this unsupervised learning paradigm toward data-efficient settings, particularly zero-shot~\cite{xu2026mrad,fang2025af,cao2024adaclip} and few-shot AD~\cite{qu2025dictas,zhang2025logsad,jeong2023winclip}, depending on the availability of normal reference samples.

% P2: existing efforts & their limitations: CLIP-based and MLLM-based
Benefiting from the strong zero/few-shot recognition capabilities of pre-trained vision-language models (VLMs), a line of work has emerged to adapt their unified visual-text representations for AD via parameter-efficient tuning, achieving impressive performance across multiple benchmarks~\cite{zhou2024anomalyclip,zhu2025fine,ma2025aa}. 
Early studies such as WinCLIP~\cite{jeong2023winclip} establish a training-free paradigm by exploiting handcrafted text prompts and multi-scale visual features from CLIP~\cite{zheng2021generative} for anomaly scoring, providing strong baselines for zero- and few-shot AD. 
\begin{wrapfigure}{r}{0.6\columnwidth}
\vspace{-10pt}
\centering
\includegraphics[width=0.55\columnwidth]{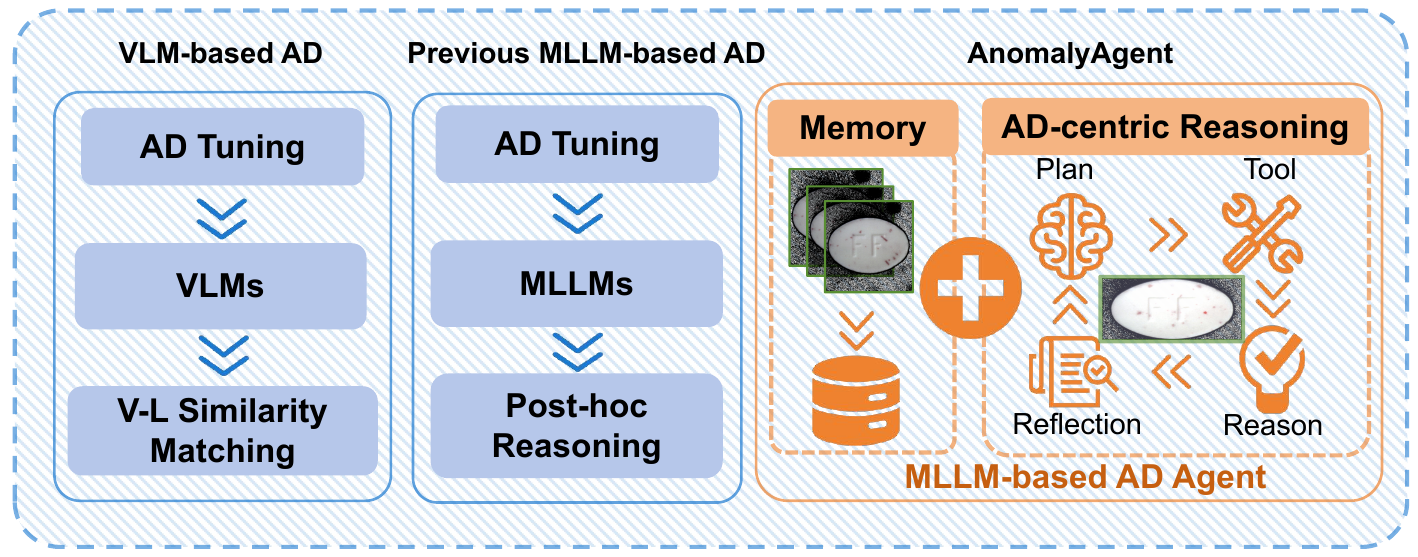}
\vspace{-3pt}
\caption{\small Comparison of conventional VLM-based AD, previous MLLM-based AD, and our AnomalyAgent. Existing VLM-based methods mainly rely on similarity-based anomaly scoring, while previous MLLM-based methods use MLLMs as post-hoc reasoning modules after task-specific adaptation. In contrast, AnomalyAgent customizes the agentic abilities of MLLMs for AD and directly integrate them into the AD process in a fully training-free manner.}
\label{fig:intro}
\vspace{-12pt}
\end{wrapfigure}
Following this direction, zero-shot AD methods typically enhance visual–text alignment via prompt learning to capture anomaly-relevant semantics~\cite{zhu2025fine, zhou2024anomalyclip, cao2024adaclip, ma2025aa}, while few-shot AD methods leverage normal reference samples to characterize expected normal patterns and detect anomalies through reference-based comparison with query samples~\cite{zhu2024inctrl, chen2023april, qu2025dictas}.

However, existing VLM-based methods still require substantial training on auxiliary datasets to adapt the VLMs to AD. Furthermore, their inference largely rely on visual–text representation similarity-based anomaly scoring, which provides only shallow alignment between text prompts and visual features, with anomalies identified via thresholding of the resulting scores~\cite{xu2026mrad,roth2022patchcore}. While effective for relatively simple and well-defined anomaly patterns, such as surface defects or localized lesions, this paradigm lacks explicit reasoning, making it insufficient for handling anomalies that require contextual, relational, or task-specific understanding. For example, detection of logical anomalies may depend on object arrangement or operational rules~\cite{bergmann2022MVTecloco,zhang2025logsad}, while real-world industrial and logistics scenarios often exhibit open-ended variations in object appearance, pose and environmental conditions~\cite{hofer2025kaputt}.

These challenges go beyond shallow representation matching and require more structured reasoning about such contextual and logical abnormalities.
Recent advances in multimodal large language models (MLLMs) offer a promising foundation for such reasoning and beyond~\cite{bai2023qwen,liu2024deepseek,li2024llava}. However, their application to the AD task remains largely underexplored. 
Existing MLLM-based AD methods typically use reasoning either for anomaly interpretation~\cite{jiang2025mmad,xu2025anomalyOV,chao2025anomalyr1, gu2023anomalyagpt}, or as auxiliary prior extraction for downstream anomaly analysis~\cite{zhang2025logsad}. In both cases, MLLM reasoning is not directly integrated into the detection decision itself, restricting its utility for zero-shot and few-shot AD. As a consequence, their effectiveness in zero-shot and few-shot anomaly detection settings remains limited, often underperforming strong VLM-based training-free baselines or requiring dataset-specific tuning to achieve competitive performance.
% Moreover, some recent efforts shift focus toward qualitative reasoning demonstrations when quantitative performance is limited, which does not fully establish their effectiveness for anomaly detection.

% \begin{figure}[!t]
% \centering
% \includegraphics[trim=0cm 0cm 0cm 0cm, clip, width=\columnwidth]{figs/intro_agent.pdf}
% \caption{Comparison of conventional VLM-based AD, previous MLLM-based AD, and the proposed AnomalyAgent. Existing VLM-based methods mainly rely on similarity-based anomaly scoring, while previous MLLM-based methods use MLLMs as post-hoc reasoning modules after task-specific adaptation. In contrast, AnomalyAgent customizes superior agentic abilities of MLLMs for AD and directly integrate them into the anomaly detection process in a fully training-free manner.}
% \label{fig:intro}
% \end{figure}

% P3: what's in need && our methods
% To tackle these limitations, we propose AnomalyAgent, a novel training-free, agentic framework that leverages the advanced reasoning and generalization capabilities of multimodal large language models (MLLMs) for zero/few-shot anomaly detection. 
% Instead of relying on fixed prompts or static similarity measures, AnomalyAgent formulates anomaly detection as a dynamic reasoning process driven by an MLLM-based agent. To this end, we first design a comprehensive anomaly-centric toolset that enables the agent to adaptively analyze visual inputs, decompose anomaly cues, and perform structured reasoning in zero-shot settings. To be specific, ...

To address these challenges, we propose \textbf{AnomalyAgent}, a fully training-free agentic framework that directly leverages the superior agentic abilities of MLLMs—planning, tool use, reasoning, and reflecting—specifically for anomaly detection. As illustrated in Fig.~\ref{fig:intro}, unlike existing VLM-based methods that rely on shallow similarity matching with auxiliary-domain tuning, or previous MLLM-based approaches that require task-specific adaptation for post-hoc reasoning, AnomalyAgent directly integrates anomaly-centric agentic abilities into the detection process itself in a fully training-free manner.
To this end, an \textbf{anomaly-centric toolset} is introduced in AnomalyAgent to progressively gather and verify visual evidence through hypothesis-driven exploration, counterfactual verification, and reflective reasoning tailored for AD. This allows it to handle complex abnormality patterns beyond simple appearance deviations.
% AnomalyAgent supports both zero-shot and few-shot anomaly detection within a unified framework. 
% In the zero-shot setting, 
% It leverages an \textbf{anomaly-centric toolset} to progressively gather and verify visual evidence. 
% Specifically, the reasoning capability of MLLMs is leveraged through tool-guided inference, allowing the model to formulate intermediate hypotheses, retrieve targeted visual evidence, verify them against expected normal patterns, and refine its predictions accordingly.
% it formulates detection as a cyclic \textbf{planning--reasoning--reflecting} process. It adaptively invokes an \textbf{anomaly-centric toolset}, including generic visual tools and cross-modal counterfactual analyzers, to inspect suspicious regions, compare observations with expected normal conditions, and refine its judgment through iterative reflection. 
In scenarios where in-context examples are available, \eg, few-shot AD settings, a \textbf{self calibration-based memory mechanism} is further introduced to convert the few-shot examples into actionable in-context memories to enable self calibration of the reasoning against dataset-specific normal patterns in AnomalyAgent. 

Overall, our contributions are summarized as follows:
\begin{itemize}
    \item We introduce \textbf{AnomalyAgent}, a novel training-free agentic framework tailored for AD. It shifts anomaly detection from visual-text representation similarity-based anomaly scoring to tool-and-memory-augmented anomaly reasoning,enabling score-thresholding-free anomaly decision and in-depth reasoning over complex abnormality patterns.
    \item We further propose an \textbf{anomaly-centric reasoning toolset} and a \textbf{self calibration-based memory mechanism} to build an agentic AD-oriented planning--reasoning--reflecting pipeline. With this pipeline, AnomalyAgent can actively gather, verify, and refine evidence against expected normal patterns throughout its reasoning, and bring in external memories to perform self calibration when such information is available.
    % . Built upon an \textbf{anomaly-centric reasoning toolset}, AnomalyAgent conducts comprehensive and in-depth analysis by integrating generic visual tools with cross-modal counterfactual analyzers. We further introduce a \textbf{twice-forwarding memory mechanism} that converts few-shot normal references into in-context normality priors without any parameter update.
    \item We establish a comprehensive evaluation benchmark covering a wide range of anomaly types across multiple AD scenarios, including industrial inspection, medical imaging, and logistics. Extensive comparisons with VLM-based methods and MLLM agents demonstrate the strong generalization ability of AnomalyAgent in both zero-shot and few-shot settings.
\end{itemize}

\section{Related Work}
\label{sec:related_work}

\noindent\textbf{VLM-based Zero-/Few-Shot Anomaly Detection.}
% Point: lacking generalization: 1) limited performance on diverse datasets, such as logical anomaly detection and assembly-line datasets, 2) relies on thresholds to differentiate anomalous samples and nominal samples.  or with only a few normal reference samples
% Traditional anomaly detection (AD) methods typically rely on application-specific datasets for model training~\cite{}. 
To improve generalization to unseen domains, zero- and few-shot anomaly detection (ZSAD/FSAD) methods have emerged to identify anomalies without dataset-specific training. Recent advances are largely driven by vision-language models (VLMs), which enable AD through visual--text similarity matching in a unified embedding space. Early methods such as WinCLIP~\cite{jeong2023winclip} establish a training-free paradigm by exploiting CLIP~\cite{radford2021CLIP} features with handcrafted prompts under both ZSAD and FSAD settings. For ZSAD, subsequent approaches mainly improve anomaly-sensitive representation learning through prompt-based adaptation~\cite{chen2023april,zhu2025fine,zhou2024anomalyclip,cao2024adaclip,gu2024filo}. For example, APRIL-GAN~\cite{chen2023april} introduces additional learnable layers trained with auxiliary anomaly data, while methods such as AnomalyCLIP~\cite{zhou2024anomalyclip}, FAPrompt~\cite{zhu2025fine}, and AdaCLIP~\cite{cao2024adaclip} employ learnable prompts to better align visual features with anomaly-relevant semantics. For FSAD, existing methods leverage normal reference samples through reference-based comparison~\cite{li2024promptad,zhu2024inctrl,ResAD}. PromptAD~\cite{li2024promptad} learns class-specific normal patterns via one-class prompt learning, while InCTRL~\cite{zhu2024inctrl} and ResAD~\cite{ResAD} model residual discrepancies between query samples and normal references for anomaly detection. However, these methods still rely on similarity-based decision mechanisms, limiting their effectiveness in scenarios requiring contextual or relational understanding.

\noindent\textbf{MLLM-based Anomaly Detection and Reasoning.}
% Point: MMAD~\cite{jiang2025mmad}, Anomaly-OV~\cite{xu2025anomalyOV}, and AnomalyR1~\cite{chao2025anomalyr1} all focus more on reasoning instead of conventional image anomaly detection performance, which limits their practicability. LogSAD~\cite{zhang2025logsad} pays more attention to few-shot logical anomaly detection.
% \gs{we should add a short paragraph here to discuss generic agentic methods and why they cannot do well in AD.}
Motivated by the strong reasoning capability of MLLMs, recent studies explore anomaly detection from a reasoning perspective. Approaches such as MMAD~\cite{jiang2025mmad}, AnomalyGPT~\cite{gu2023anomalyagpt}, Anomaly-OV~\cite{xu2025anomalyOV}, and AnomalyR1~\cite{chao2025anomalyr1} mainly treat MLLMs as post-hoc reasoning modules for anomaly analysis through natural language descriptions or multi-step inference. Other methods, such as LogSAD~\cite{zhang2025logsad}, leverage MLLMs to model structural or logical constraints as prior knowledge for AD. However, existing methods still lack anomaly-aware reasoning guidance that explicitly integrates reasoning into the anomaly detection process itself, often requiring dataset-specific tuning and underperforming strong training-free VLM baselines.
Recent agentic frameworks have demonstrated strong capabilities in planning, tool use, and multi-step reasoning across multimodal tasks~\cite{yao2023react,shinn2023reflexion,chhikara2025mem0}. However, directly applying generic agents to AD remains challenging due to the lack of anomaly-specific reasoning guidance and calibrated normality priors. In contrast, AnomalyAgent is a fully training-free agentic AD framework that explicitly integrates MLLM reasoning into anomaly detection through an anomaly-centric reasoning toolset and a self-calibration-based memory mechanism.

\section{Methodology}
\label{sec:method}

\noindent\textbf{Problem Statement.}
Existing ZSAD/FSAD methods~\cite{zhu2025fine,zhou2024anomalyclip,zhu2024inctrl} typically require auxiliary datasets to adapt VLMs to AD. Specifically, given auxiliary data $\mathcal{D}_{\text{aux}} = \{X_{\text{train}}, Y_{\text{train}}\}$, where $X_{\text{train}} = \{x_i\}_{i=1}^{N}$ denotes the training images, $Y_{\text{train}} = \{y_i\}_{i=1}^{N}$ denotes the corresponding binary labels, and each sample $x_i$ satisfies $y_i \in \{0, 1\}$ with $0$ and $1$ represent normal an anomalous samples respectively, a pretrained VLM-based anomaly scoring function $f_{\text{VLM}}(x_i, \text{Prompt})$ is fine-tuned using $\mathcal{D}_{\text{aux}}$ and then directly applied to yield an \textbf{anomaly score} for samples from unseen test data $\mathcal{D}_{\text{test}}$. 
In contrast, our work operates in a \textbf{fully training-free} manner, requiring neither auxiliary data training nor parameter adaptation.
In particular, our goal is to develop a generalizable MLLM-driven anomaly reasoning agent $f_{\text{
MLLM}}(x_i, \text{Prompt}, \text{Tool}, \text{Memory})$ that can effectively generalize across different unseen test datasets $\mathcal{T} = \{\mathcal{D}^1_{\text{test}}, \mathcal{D}^2_{\text{test}}, \cdots, \mathcal{D}^m_{\text{test}}\}$ under both zero-shot and few-shot settings, where each test dataset $\mathcal{D}^i_{\text{test}} = \{X^i_{\text{test}}, Y^i_{\text{test}}\}$ is drawn from a distinct application domain and $f_{\text{
MLLM}}$ does not require any training. 
% The anomaly distributions in these domains differ from those observed in the auxiliary training dataset $\mathcal{D}$.
In the zero-shot setting, no target-domain reference samples are available during inference. In contrast, FSAD additionally provides a small set of normal reference samples from the target domain, denoted as $\mathcal{P} = \{p_1, p_2, ..., p_k\}$, where $k$ is typically a small number (\eg, $k \ll N$). 
% $\mathcal{P}$ is only available during inference and are not used for parameter updating or model adaptation.
% 这里confirm一下
The resulting model is expected to directly produce a \textbf{hard-label prediction}, \textit{ie}, ``anomalous'' or ``normal'', providing a global prediction of
whether the query image contains any abnormal content.

\begin{wrapfigure}[14]{r}{0.65\columnwidth}
\vspace{-10pt}
\centering
\includegraphics[
trim=0cm 0cm 0cm 0cm,
clip,
width=0.6\columnwidth
]{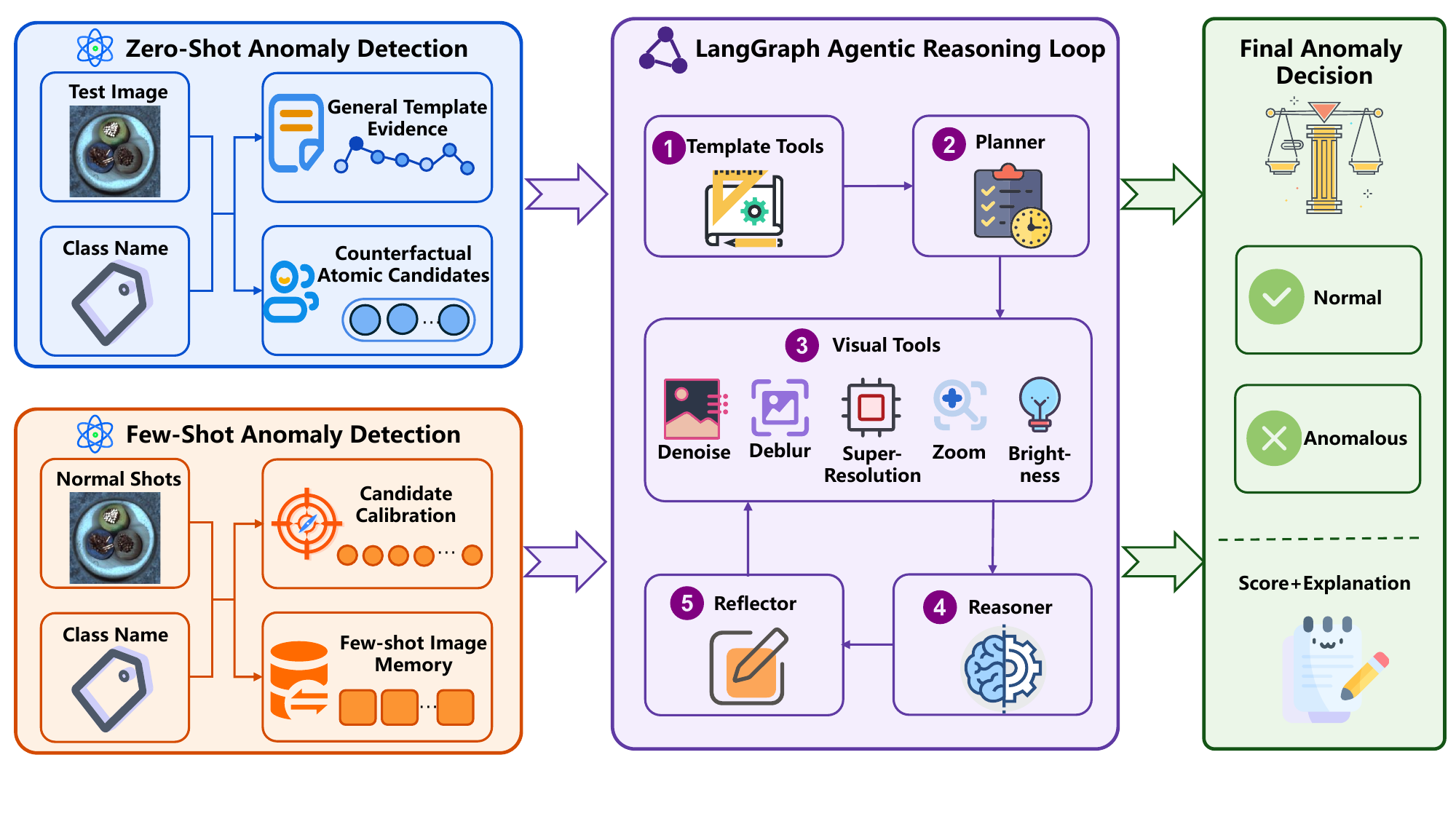}
\vspace{-10pt}
\caption{\small Overview of AnomalyAgent.}
\label{fig:framework}
\vspace{-10pt}
\end{wrapfigure}

\noindent\textbf{Overview of AnomalyAgent.}
% In this work, we propose \textbf{AnomalyAgent}, a novel training-free, agentic framework that leverages the advanced reasoning and generalization capabilities of multimodal large language models (MLLMs) for anomaly detection. 
As illustrated in Fig.~\ref{fig:framework}, AnomalyAgent inputs the image and the class name into the template tools, deriving the general template analysis and class-specific counterfactual template analysis. Given the general template analysis and the raw image, the Planner selects and invokes an appropriate visual tool to enhance anomaly-relevant evidence. Then, the augmented image, raw image, and template analysis are fed into the Reasoner for judgment. If the termination condition is not met, the aforementioned information, as well as the Reasoner's thought, will be input into the Reflector for a new loop of investigation.
Below, we present these modules in detail.
% \begin{figure}[!t]
% \centering
% \includegraphics[trim=0cm 0cm 0cm 0cm, clip, width=0.85\columnwidth]{figs/AgentAD_diagram.pdf}
% \caption{In zero-shot and few-shot AD, existing score-based methods rely on manual thresholds, while generic agent or memory-based methods may fail to verify visual evidence or over-trust stored normal patterns. AnomalyAgent addresses these limitations by coupling visual tool use with prior- and memory-guided reflection, enabling more reliable and explainable anomaly decisions.}
% \label{fig:framework}
% \end{figure}

% Polish Plan
% In 3.1 ZSAD
% \paragraph{Agent Framework}
% \paragraph{Anomaly-centric Reasoning Toolset}

% In 3.2 Few-shot Anomaly Detection
% \paragraph{Tool Calibration}
% \paragraph{Twice-forwarding Memory Integration}

% \subsection{Anomaly-centric Reasoning Toolset}
% \subsubsection{Semantic Prior Construction}
% \subsubsection{Counterfactual Evidence Generation}
% \subsubsection{Planning-Reasoning-Reflection}

% \subsection{Self Calibration-based Memory for Few-shot Anomaly Detection}
% \subsubsection{Memory Construction}
% \subsubsection{}

% \subsection{Zero-shot Anomaly Detection} %GS: Tool set introduced in this part is also used in the few-shot setting, so it's not accurate to name it as ZSAD

\subsection{Anomaly-Centric Toolset for AD-Oriented Agentic Planning--Reasoning--Reflecting}

% To achieve anomaly detection under a training-free setting, we design a zero-shot agentic anomaly detection framework that can adapt to different inspection requirements without using any training samples. Given an input image $x$ and its semantic class name $c$, the objective is to determine whether $x$ contains abnormal visual evidence with respect to the expected appearance of class $c$. Different from conventional anomaly detection methods that rely on learned normal distributions, our framework constructs task-conditioned semantic priors from language and performs progressive visual reasoning with a multimodal agent.

% Technically, the proposed zero-shot pipeline consists of three key stages: (1) class-conditioned semantic prior construction, (2) counterfactual evidence generation, and (3) agentic visual reasoning with self-reflection.
To enable anomaly detection in a fully training-free setting, we present a zero-shot agentic framework that can adapt to diverse inspection requirements without accessing training samples or performing parameter updates. Unlike generic MLLM agents that mainly rely on open-ended visual question answering, AnomalyAgent is specifically designed for anomaly detection from two complementary aspects: an \textbf{anomaly-centric toolset} that provides task-relevant visual and semantic evidence, and an  \textbf{AD-oriented agentic pipeline} that progressively plans, reasons, and reflects over such evidence before making a hard-label prediction.

\paragraph{Anomaly-Centric Toolset.}
The tools in AnomalyAgent are organized into two categories. The first category consists of \textbf{general visual tools}, which enhance the query image or focus on local regions to provide clearer visual evidence for anomaly 
inspection. The second category consists of \textbf{template tools}, which exploit both textual priors and image-level descriptions to construct structured anomaly-related cues. This design enables the agent to improve the visibility of suspicious observations while grounding its reasoning in explicit AD-specific priors.

\textbf{General Visual Tools.}
We include five general visual tools: image denoising, image deblurring, super-resolution, zooming, and brightness adjustment. These tools are invoked when the Planner or the Reflector identifies insufficient or ambiguous visual evidence, such as low image quality, small suspicious regions, blur, or poor illumination. The implementations are provided in the Appendix~\ref{app:tool_details}.

\textbf{Template Tools.} include a general template tool and a counterfactual atomic template tool. The \textbf{general template tool} is inspired by the template ensembles in WinCLIP~\cite{jeong2023winclip}, \ie, normal ensembles ($E_N$) and anomaly ensembles ($E_A$). However, instead of restricting anomaly evidence to the matching between CLIP visual features and predefined textual templates, we leverage the fine-grained visual description capability of MLLMs. Specifically, an MLLM-based image captioning model is employed to generate detailed descriptions of the query image $x$ from complementary visual perspectives:
\begin{equation}
    \text{Captions} = \{d_1,d_2,d_3\} = \text{MLLM}(x; \text{Prompt}_\text{IC}),
\end{equation}
where each caption $d_i$ focuses on a different visual aspect of the object and the $\text{Prompt}_\text{IC}$ can be found in the Appendix~\ref{app:prompts}.

These captions are then compared with class-conditioned normal and anomaly templates, yielding a preliminary semantic evidence report that indicates whether the observed appearance is more consistent with normality or abnormality:
\begin{equation}
    \text{Report}_\text{GR} = \text{MLLM}(\text{Emb}(\text{Captions}) \times E_A, \text{Emb}(\text{Captions}) \times E_N),
\end{equation}
where $\text{Emb}(\cdot)$ stands for an text embedding model.

The \textbf{counterfactual atomic template tool} works in a similar way while being more adaptive and flexible. An LLM is utilized to generate class-conditioned anomalous and normal candidate sets:
\begin{equation}
    E^C_A; E^C_N = \text{LLM}(c; \text{Prompt}_\text{CG}).
\end{equation}
The anomaly candidates describe concrete abnormal patterns such as surface damage, deformation, missing parts, contamination, or structural defects, while the normal candidates describe intact and regular appearances. 
For each caption $d_j$, we compute its similarity to the normal and anomalous semantic prototypes, yielding a soft anomaly score:
\[
s_j = \sigma \left(
\mathrm{sim}(\text{Emb}({d_j}), p_A) -
\mathrm{sim}(\text{Emb}(d_j), p_N)
\right),
\]
where $p_A$ and $p_N$ are the averaged embeddings of anomaly and normal textual prototypes. This produces an interpretable preliminary evidence report, including caption-level anomaly scores, similarity margins, and agreement patterns across perspectives.
% \textbf{Target-specified Anomaly Prior.}
% In some zero-shot settings, the task does not ask for general anomaly detection, but instead specifies which anomaly type should be regarded as positive. To handle this setting, we condition the semantic prior on a target anomaly description $t$. The candidate generation process becomes:
% \[
% \mathcal{A}_{c,t}, \mathcal{N}_{c,t}
% =\mathrm{LLM}(c,t,\mathrm{Prompt}_{target}),
% \]
% where $\mathcal{A}_{c,t}$ contains only target-positive descriptions that satisfy the specified anomaly definition, while $\mathcal{N}_{c,t}$ contains target-negative descriptions. Importantly, target-negative cases may include non-target visual irregularities. Therefore, the model is prevented from expanding the task into generic anomaly detection. During reasoning, the same target description is also injected into the agent prompt, so that the planner, reasoner, and reflector consistently distinguish target-positive evidence from irrelevant non-target variations.
In addition, we further construct a counterfactual evidence report through atomic candidate matching. Specifically, each image caption is compared against the generated anomaly and normal candidate sets:
\begin{equation}
\mathrm{TopK}_A(d_j), \mathrm{TopK}_N(d_j)
=
\mathrm{RetrieveTopK}
\left(
\text{Emb}(d_j), \text{Emb}(E^C_A), \text{Emb}(E^C_N)
\right),
\end{equation}
where $K$ is a hyperparameter. The retrieved candidates provide fine-grained evidence explaining whether the visual description is closer to abnormal or normal semantic patterns. We compute the candidate-level margin:
\begin{equation}
m_j =
\max_{a_i \in E^C_A} \mathrm{sim}(d_j,a_i)
-
\max_{n_i \in E^C_N} \mathrm{sim}(d_j,n_i),
\end{equation}
and summarize the top-matched anomaly and normal candidates as counterfactual evidence:
\begin{equation}
    \text{Report}_\text{CR} = \text{LLM}(s_1, s_2, s_3, \mathrm{TopK}_A(d_j), \mathrm{TopK}_N(d_j), m_1,m_2,m_3).
\end{equation}
This report is not directly used as a hard decision rule; instead, it provides structured semantic cues for the agent to reason over.

% For logical anomaly detection, where abnormality is often caused by violations of quantity, spatial relation, ordering, or structural consistency, appearance-level textual matching is insufficient. Therefore, we introduce an additional counterfactual template for rule-level verification. This template decomposes logical normality assumptions into atomic visual checks:
% \[
% \mathcal{R}_c=\{r_k\}_{k=1}^{K},
% \]
% where each $r_k$ is a visually verifiable condition. The VLM evaluates each check against the image and returns a status from $\{\mathrm{pass}, \mathrm{fail}, \mathrm{unknown}\}$. High-confidence failures are treated as hard rule-violation evidence, while unknown checks indicate insufficient visual support rather than normality. In this way, logical anomaly detection is handled through explicit counterfactual verification instead of dataset-specific rules.

\textbf{AD-Oriented Agentic Pipeline.}
After obtaining $\text{Report}_\text{GR}$ and $\text{Report}_\text{CR}$, AnomalyAgent performs anomaly judgment through a planning--reasoning--reflecting pipeline. The \textbf{Planner} first receives the raw image, class name, and semantic evidence reports, and converts them into an inspection plan:
% \begin{equation}
%     \text{Plan} =
%     \text{MLLM}_\text{P}
%     (x,c,\text{Report}_\text{GR},\text{Report}_\text{CR};\text{Prompt}_\text{P})
%     \rightarrow
%     \{\text{Potential Anomalies},\text{Heuristic Prompt},\text{Tools to Use}\}.
% \end{equation}
\begin{equation}
\begin{aligned}
    \text{Plan} &=
    \text{MLLM}_\text{P}
    (x,c,\text{Report}_\text{GR},\text{Report}_\text{CR};\text{Prompt}_\text{P}) \\
    &\rightarrow
    \{\text{Potential Anomalies},\text{Heuristic Prompt},\text{Tools to Use}\}.
\end{aligned}
\end{equation}
Here, the potential anomalies summarize suspicious visual regions or abnormal patterns suggested by the template reports, while the heuristic prompt provides class-specific inspection guidance for subsequent reasoning. When the visual evidence is insufficient, the Planner invokes the corresponding general visual tools to obtain enhanced observations $\tilde{x}$.

The \textbf{Reasoner} then integrates the raw image, enhanced observations, inspection plan, and semantic evidence reports to produce a structured anomaly judgment:
\begin{equation}
\begin{aligned}
    \text{Result}
    &= \text{MLLM}_\text{R}
    \bigl(x,\tilde{x},c,\text{Plan}, \\
    &\quad \text{Report}_\text{GR},\text{Report}_\text{CR};
    \text{Prompt}_\text{R}\bigr)
    \rightarrow
    \{\hat{y},\text{Reason}\}.
\end{aligned}
\end{equation}
The prediction $\hat{y}$ is selected from $\{\text{normal},\text{anomalous},\text{uncertain}\}$, where $\text{Reason}$ explains the supporting visual observations and their consistency with the semantic priors. If a confident normal or anomalous decision is obtained, the pipeline directly returns the result.

When the Reasoner outputs $\text{uncertain}$, the \textbf{Reflector} revisits the previous plan and judgment, identifies missing or ambiguous evidence, and refines the inspection instruction:
% \begin{equation}
%     \text{Reflect} =
%     \text{MLLM}_\text{Ref}
%     (x,\tilde{x},c,\text{Plan},\text{Result},
%     \text{Report}_\text{GR},\text{Report}_\text{CR};
%     \text{Prompt}_\text{Ref})
%     \rightarrow
%     \{\text{Missing Evidence},\text{Ambiguous Evidence},
%     \text{Refined Heuristic Prompt},\text{Tools to Use}\}.
% \end{equation}
\begin{equation}
\begin{aligned}
    \text{Reflect} &=
    \text{MLLM}_\text{Ref}
    (x,\tilde{x},c,\text{Plan},\text{Result},
    \text{Report}_\text{GR},\text{Report}_\text{CR};
    \text{Prompt}_\text{Ref}) \\
    &\rightarrow
    \{\text{Missing Evidence},\text{Ambiguous Evidence},
    \text{Refined Heuristic Prompt},\text{Tools to Use}\}.
\end{aligned}
\end{equation}
The reflected instruction is then sent back to the Planner or Reasoner for targeted re-inspection. In this way, AnomalyAgent avoids premature decisions from incomplete evidence while keeping the final prediction as a hard-label anomaly decision.

% \subsection{Memory Agent for Few-Shot Anomaly Detection}
\subsection{Memory-Augmented Self Calibration with Normal Reference Examples}
\label{sec:memory_agent}
\vspace{-0.5em}

To further reduce false alarms caused by class-specific normal variations, we introduce a memory-augmented self-calibration mechanism that uses a small set of normal reference examples without updating any model parameters. For each class $c$, let
\begin{equation}
    \mathcal{M}_c^N=\{x_i^N\}_{i=1}^{K}
\end{equation}
denote the available normal reference images. We first run the same AnomalyAgent pipeline on each $x_i^N$ and record its intermediate evidence:
\begin{equation}
    \mathcal{R}_i =
    \{\text{Captions}_i,\text{Report}_{\text{GR}}^i,
    \text{Report}_{\text{CR}}^i,\hat{y}_i,\text{Reason}_i\}.
\end{equation}
Since all $x_i^N$ are known to be normal, these records provide self-calibration signals for identifying normal visual patterns that may be mistakenly treated as anomalies.

\paragraph{Counterfactual candidate calibration.}
For each atomic candidate $e\in E_A^C\cup E_N^C$, we assign a reliability weight $w_e$, initialized to $0.5$. Given the captions of a normal reference image, an LLM judges the relation between the reference description and each candidate as $\{\text{fit},\text{conflict},\text{unrelated}\}$. We compute the caption-candidate matching strength
\begin{equation}
    \rho_i(e)=\max_j \mathrm{sim}\bigl(\text{Emb}(d_{ij}),\text{Emb}(e)\bigr).
\end{equation}
The weight is then updated according to the normal-reference constraint:
\begin{equation}
w_e \leftarrow \mathrm{clip}(w_e+\Delta_i(e),0,1),
\end{equation}
where normal candidates are strengthened when they fit normal references, while anomaly candidates are weakened when they fit normal references. In practice, if the original counterfactual evidence incorrectly favors anomaly on a normal reference image, a larger update coefficient is used. The calibrated weights are then injected into atomic candidate matching:
\begin{equation}
    \ell(d_j,e)=
    \frac{\mathrm{sim}(\text{Emb}(d_j),\text{Emb}(e))}{\tau_{\text{cand}}}
    +(w_e-0.5),
\end{equation}
where $w_e=0.5$ recovers the original unweighted matching. The calibrated counterfactual margin is computed as
\begin{equation}
    m_j^{\text{cal}}=
    \tau_{\text{cand}}\max_{a\in E_A^C}\ell(d_j,a)
    -
    \tau_{\text{cand}}\max_{n\in E_N^C}\ell(d_j,n),
\end{equation}
which produces a refined $\text{Report}_{\text{CR}}$ for subsequent agent reasoning.

\paragraph{Normal-reference memory.}
In addition to candidate calibration, we build a normal-reference memory bank. Each reference image is stored with its CLIP image embedding, captions, template margins, final judgment, and reasoning trace:
\begin{equation}
    \mathcal{B}_c =
    \{(\phi(x_i^N),\mathcal{R}_i)\}_{i=1}^{K},
\end{equation}
where $\phi(\cdot)$ denotes the image-level embedding. For normal references that are incorrectly judged as anomalous, the agent further reflects on the failure case and extracts class-level hard-normal cues, \ie, suspicious-looking but normal visual patterns. These cues are summarized into a class calibration note $H_c$ and enabled only when a validation pass on normal references improves correctness without degrading previously correct cases.

During inference, the query image $x$ retrieves visually similar normal memories:
\begin{equation}
    \mathcal{N}_\gamma(x)=
    \{(x_i^N,\mathcal{R}_i)\mid
    \cos(\phi(x),\phi(x_i^N))>\gamma\},
\end{equation}
where $\gamma$ is a similarity threshold. The retrieved entries are summarized into a memory report $\text{Report}_{\text{MEM}}$, including similarity statistics, average template margins, and the reliability of previous judgments on retrieved normal references.

Finally, the Reasoner receives both semantic reports and memory-based calibration cues:
\begin{equation}
\begin{aligned}
    \text{Result}
    &= \text{MLLM}_{\text{R}}
    \bigl(x,\tilde{x},c,\text{Plan},
    \text{Report}_{\text{GR}},\text{Report}_{\text{CR}}^{\text{cal}}, \\
    &\quad \text{Report}_{\text{MEM}}, H_c;
    \text{Prompt}_{\text{R}}\bigr)
    \rightarrow
    \{\hat{y},\text{Reason}\}.
\end{aligned}
\end{equation}
The memory report is used only as historical normal-shot calibration rather than a hard decision rule. Thus, the final prediction remains a hard-label anomaly judgment, while the agent is calibrated against class-specific normal variations and hard-negative normal appearances.

\section{Experiments}
\label{sec:exp}

\paragraph{Datasets.} 
To verify the effectiveness of AnomalyAgent, we conduct extensive experiments across five real-world datasets from diverse application domains, including industrial defect dataset (MVTec~\cite{bergmann2019MVTec}), medical image datasets (HeadCT~\cite{salehi2021multiresolution} and LAG~\cite{li2019attention}), logistics defect dataset (Kaputt~\cite{hofer2025kaputt}), and logical anomaly dataset (MVTec LOCO~\cite{bergmann2022MVTecloco}). These are test datasets, which are used directly to evaluate the detection performance in the zero-shot setting. 
In the few-shot setting, the few-shot normal prompts for the target data are randomly sampled from the training set of target datasets and remain the same for all detection models for fair comparison. We evaluate the performance with varying number of few-shot normal sample set: $K = 1, 2, 4$ (See \texttt{Appendix} for details about the dataset settings).

% To cope with our motivation for developing general image detection MLLM agents, AnomalyAgent is verified on diverse datasets, including industrial products dataset,  \textit{ie}, MVTec~\cite{bergmann2019MVTec}, medical image dataset,  \textit{ie}, HeadCT~\cite{salehi2021multiresolution} and LAG~\cite{li2019attention}, large logistics dataset, \textit{ie}, Kaputt~\cite{hofer2025kaputt}, and logical anomaly detection dataset, \textit{ie}, MVTec LOCO~\cite{bergmann2022MVTecloco}. 
% These datasets are used both in ZSAD and FSAD.
% More details of these datasets can be found in the Appendix. 

\paragraph{Baseline Methods and Evaluation Metrics.}
We compare AnomalyAgent with several state-of-the-art (SotA) anomaly detection methods under both zero-shot and few-shot settings. For ZSAD, the competing methods include training-free baselines, \textit{ie}, CLIP~\cite{radford2021CLIP}, WinCLIP~\cite{jeong2023winclip}, MRAD~\cite{xu2026mrad}, and ReAct Agent~\cite{yao2023react}. We also include trainable ZSAD approaches, including APRIL-GAN~\cite{chen2023april}, AnomalyCLIP~\cite{zhou2024anomalyclip}, and FAPrompt~\cite{zhu2025fine}, but they are not included for direct performance comparison; their results are only used to check the performance gap between training-free and trainable AD methods. For FSAD, the training-free baselines include WinCLIP+~\cite{jeong2023winclip}, PatchCore~\cite{roth2022patchcore}, and an agentic method Mem0~\cite{chhikara2025mem0}, while the trainable baselines include APRIL-GAN~\cite{chen2023april} and InCTRL~\cite{zhu2024inctrl}. 
% Since AnomalyAgent is the first agentic framework tailored for AD, we additionally re-implement the generic ReAct agent~\cite{yao2023react} as an MLLM-based ZSAD baseline. Furthermore, to evaluate the effectiveness of the proposed memory mechanism in FSAD, we integrate Mem0~\cite{chhikara2025mem0} with our anomaly-centric reasoning toolset as an MLLM-based FSAD baseline for comparison. 
Note that ReAct is not included in the FSAD experiments as it cannot directly use the few-shot references; similarly, Mem0 that requires few-shot references is not used in the ZSAD experiments.
Following previous AD approaches~\cite{cao2024adaclip,zhang2025logsad,fuvcka2026anomalyvfm}, we report the Area Under the Receiver Operating Characteristic (AUROC) and the highest possible F1 score (F1-max) as our evaluation metrics.

\paragraph{Implementation Details.}
% We implemented our method in Python using LangGraph and LangChain. 
We use GPT-5.1~\cite{openai2025gpt51} with the Response API as the default MLLM for all agentic methods (AnomalyAgent, ReACT, and Mem0) for image description, planning, reasoning, and reflection in the main experiments. To reduce inference cost, GPT-4.1-mini~\cite{openai2025gpt41} is used for auxiliary reasoning and memory-related operations.
% % The maximum numbers of tool calls and reasoner iterations are set to 5 and 6, respectively. To improve decision reliability, a sample is predicted as normal only after two consecutive normal judgments.
For embedding-based evidence matching, we employ a Qwen3-4B-based text embedding model~\cite{yang2025qwen3} by default. 
For training-free methods, all datasets are directly used for inference without any training or adaptation. For trainable methods, we follow the official experimental protocols of the competing approaches, where models are trained on their corresponding auxiliary datasets and evaluated on the target test sets without further fine-tuning. 
Implementation details for AnomalyAgent and the competing methods are provided in \texttt{Appendix}.
% The sigmoid temperature for caption--prototype scoring is set to 0.3, while the log-sum-exp temperature for atomic candidate matching is set to 0.1.

\subsection{Main Results}
% This part consists of the experimental results of image-level zero-shot and few-shot anomaly detection.

\noindent\textbf{Zero-shot Anomaly Detection.}
Table~\ref{zero-shot} reports the ZSAD results across five datasets for both fully training-free and trainable methods. Overall, AnomalyAgent achieves the best average performance among all training-free methods, surpassing the strongest training-free baseline by up to 2.7\% AUROC and 1.7\% $F_{1}$-$\max$. Compared with conventional VLM-based approaches relying on visual--language similarity matching, AnomalyAgent shows particularly strong improvements on datasets requiring contextual and relational understanding, such as MVTec LOCO and Kaputt, and even achieves competitive performance against several trainable methods. Although trainable approaches such as AnomalyCLIP and FAPrompt still obtain slightly better results on some datasets due to auxiliary-domain optimization, AnomalyAgent remains highly competitive without any training or parameter adaptation. These results demonstrate the effectiveness of the proposed agentic reasoning framework for handling contextual, structural, and task-dependent anomalies.

As for the generic MLLM agent ReAct, it achieves competitive performance compared with WinCLIP, demonstrating the potential of agentic reasoning for AD. However, a clear gap still exists between ReAct and AnomalyAgent, suggesting that generic MLLM agents lack AD-specific reasoning guidance and calibrated normality priors for reliable anomaly understanding. In contrast, AnomalyAgent explicitly organizes anomaly-centric tools within a structured reasoning workflow, enabling more reliable anomaly decisions across diverse AD scenarios. Moreover, the consistently stronger $F_{1}$-$\max$ results achieved by MLLM-based methods, particularly ReAct and AnomalyAgent, indicate that reasoning-based anomaly analysis produces more discriminative anomaly decisions with better normal/anomaly separation than conventional anomaly scoring-based methods.

\definecolor{lightgraybg}{RGB}{245,245,245}
\definecolor{graytext}{RGB}{120,120,120}

\newcommand{\trainable}[1]{\cellcolor{lightgraybg}\textcolor{graytext}{#1}}

\definecolor{bestred}{RGB}{205,70,70}
% \definecolor{secondblue}{RGB}{65,110,185}
% \newcommand{\best}[1]{\textcolor{bestred}{#1}}
% \newcommand{\second}[1]{\textcolor{secondblue}{#1}}
\newcommand{\best}[1]{\textbf{\textcolor{bestred}{#1}}}
\newcommand{\second}[1]{\underline{#1}}

\begin{table*}[t]
\centering
\caption{
AUROC and F1-max results across five datasets under ZSAD settings.
Best results among training-free methods are highlighted in \best{red}, while second-best results are \second{underlined}.
Trainable methods are shaded in gray for reference only since they require additional training.}
\label{tab:main_results}
\setlength{\tabcolsep}{3.2pt}
\renewcommand{\arraystretch}{1.12}

\resizebox{\textwidth}{!}{
\begin{tabular}{l*{16}{c}}
\toprule

\multirow{3}{*}{\textbf{Dataset}}
& \multicolumn{10}{c}{\textbf{Training-Free Methods}}
& \multicolumn{6}{c}{\cellcolor{lightgraybg}\textcolor{graytext}{\textbf{Trainable Methods}}} \\

\cmidrule(lr){2-11}
\cmidrule(lr){12-17}

& \multicolumn{2}{c}{WinCLIP}
& \multicolumn{2}{c}{MRAD-TF}
& \multicolumn{2}{c}{ReACT}
& \multicolumn{2}{c}{\textbf{AnomalyAgent}}
& \multicolumn{2}{c}{CLIP}
& \multicolumn{2}{c}{\cellcolor{lightgraybg}\textcolor{graytext}{AprilGAN}}
& \multicolumn{2}{c}{\cellcolor{lightgraybg}\textcolor{graytext}{AnomalyCLIP}}
& \multicolumn{2}{c}{\cellcolor{lightgraybg}\textcolor{graytext}{FAPrompt}} \\

\cmidrule(lr){2-3}
\cmidrule(lr){4-5}
\cmidrule(lr){6-7}
\cmidrule(lr){8-9}
\cmidrule(lr){10-11}
\cmidrule(lr){12-13}
\cmidrule(lr){14-15}
\cmidrule(lr){16-17}

& AUROC & F1-max
& AUROC & F1-max
& AUROC & F1-max
& AUROC & F1-max
& AUROC & F1-max
& \textcolor{graytext}{AUROC} & \textcolor{graytext}{F1-max}
& \textcolor{graytext}{AUROC} & \textcolor{graytext}{F1-max}
& \textcolor{graytext}{AUROC} & \textcolor{graytext}{F1-max} \\

\midrule

MVTec
& \best{90.4} & \best{92.7}
& 81.2 & 89.3
& 74.3 & 86.3
& \second{84.5} & \second{90.7}
& 74.0 & 74.7
& \trainable{86.1} & \trainable{83.5}
& \trainable{91.6} & \trainable{89.6}
& \trainable{91.8} & \trainable{89.7} \\

MVTec LOCO
& 59.1 & \second{53.8}
& 58.2 & \best{77.3}
& \second{60.0} & \textbf{\best{77.3}}
& \best{62.8} & \best{77.3}
& 53.6 & 44.9
& \trainable{58.6} & \trainable{39.8}
& \trainable{61.9} & \trainable{57.4}
& \trainable{59.5} & \trainable{51.3} \\

LAG
& 59.6 & \second{65.4}
& \second{60.1} & \best{81.3}
& 57.7 & \best{81.3}
& \best{65.3} & \best{81.3}
& 62.6 & 66.7
& \trainable{80.7} & \trainable{77.9}
& \trainable{76.4} & \trainable{77.9}
& \trainable{74.9} & \trainable{80.2} \\

HeadCT
& \second{86.7} & 81.5
& 74.7 & 72.4
& \best{91.0} & \second{91.2}
& \best{91.0} & \best{91.5}
& 61.2 & 66.7
& \trainable{89.7} & \trainable{82.4}
& \trainable{93.4} & \trainable{89.5}
& \trainable{94.4} & \trainable{90.4} \\

Kaputt
& \second{55.6} & 36.6
& 47.6 & \second{48.3}
& 53.6 & 45.0
& \best{61.3} & \best{48.9}
& 55.0 & 41.3
& \trainable{61.4} & \trainable{42.4}
& \trainable{55.4} & \trainable{48.8}
& \trainable{63.9} & \trainable{50.6} \\

\midrule

Average
& \second{70.3} & 66.0
& 64.4 & 73.7
& 67.3 & \second{76.2}
& \best{73.0} & \best{77.9}
& 61.3 & 58.9
& \trainable{75.3} & \trainable{65.2}
& \trainable{75.7} & \trainable{72.6}
& \trainable{76.9} & \trainable{72.4} \\

\bottomrule
\end{tabular}
}
\label{zero-shot}
\vspace{-12pt}
\end{table*}

\vspace{-0.5em}
\begin{table*}[t]
\centering
\caption{
AUROC and F1-max results across five datasets under FSAD settings.
Best results among training-free methods are highlighted in \best{red}, while second-best results are \second{underlined}.
Trainable methods are shaded in gray for reference only since they require additional training.
}
\label{tab:fewshot_results}

\setlength{\tabcolsep}{3.2pt}
\renewcommand{\arraystretch}{1.12}

\resizebox{\textwidth}{!}{
\begin{tabular}{ll*{12}{c}}
\toprule

\multirow{3}{*}{\textbf{Shot}}
& \multirow{3}{*}{\textbf{Dataset}}
& \multicolumn{8}{c}{\textbf{Training-Free Methods}}
& \multicolumn{4}{c}{\cellcolor{lightgraybg}\textcolor{graytext}{\textbf{Trainable Methods}}} \\

\cmidrule(lr){3-10}
\cmidrule(lr){11-14}

&
& \multicolumn{2}{c}{WinCLIP+}
& \multicolumn{2}{c}{PatchCore}
& \multicolumn{2}{c}{Mem0}
& \multicolumn{2}{c}{\textbf{AnomalyAgent}}
& \multicolumn{2}{c}{\cellcolor{lightgraybg}\textcolor{graytext}{AprilGAN}}
& \multicolumn{2}{c}{\cellcolor{lightgraybg}\textcolor{graytext}{InCTRL}} \\

\cmidrule(lr){3-4}
\cmidrule(lr){5-6}
\cmidrule(lr){7-8}
\cmidrule(lr){9-10}
\cmidrule(lr){11-12}
\cmidrule(lr){13-14}

&
& AUROC & F1-max
& AUROC & F1-max
& AUROC & F1-max
& AUROC & F1-max
& \textcolor{graytext}{AUROC} & \textcolor{graytext}{F1-max}
& \textcolor{graytext}{AUROC} & \textcolor{graytext}{F1-max} \\

\midrule

\multirow{6}{*}{1-shot}

& MVTec
& \best{94.1} & \best{92.0}
& 83.8 & 87.4
& 83.0 & 89.2
& \second{85.0} & \second{91.0}
& \trainable{92.2} & \trainable{89.4}
& \trainable{93.0} & \trainable{91.5} \\

& MVTec LOCO
& \second{61.2} & 56.0
& 60.9 & 54.0
& 53.3 & \second{77.3}
& \best{64.0} & \best{78.6}
& \trainable{64.6} & \trainable{50.0}
& \trainable{60.0} & \trainable{57.3} \\

& LAG
& 55.1 & 50.5
& 40.8 & 49.3
& \second{60.5} & \second{81.3}
& \best{66.1} & \best{82.0}
& \trainable{77.3} & \trainable{68.8}
& \trainable{47.0} & \trainable{67.0} \\

& HeadCT
& 81.3 & 82.0
& 65.1 & 85.6
& \second{89.0} & \second{89.8}
& \best{91.1} & \best{91.2}
& \trainable{89.6} & \trainable{83.4}
& \trainable{90.2} & \trainable{91.6} \\

& Kaputt
& \second{55.9} & \second{55.6}
& 51.0 & \best{63.0}
& 55.6 & 46.3
& \best{61.9} & 49.5
& \trainable{57.4} & \trainable{56.8}
& \trainable{58.0} & \trainable{57.2} \\

\cmidrule(lr){2-14}

& Average
& \second{69.5} & 67.2
& 60.3 & 67.9
& 68.3 & \second{76.8}
& \best{73.6} & \best{78.5}
& \trainable{76.2} & \trainable{69.7}
& \trainable{69.6} & \trainable{72.9} \\

\midrule

\multirow{6}{*}{2-shot}

& MVTec
& \best{95.0} & \best{92.5}
& \second{86.8} & 88.4
& 80.6 & 89.0
& 85.6 & \second{91.2}
& \trainable{82.8} & \trainable{88.6}
& \trainable{94.3} & \trainable{92.9} \\

& MVTec LOCO
& \second{63.3} & 67.2
& 61.7 & 53.9
& 52.6 & \second{77.3}
& \best{66.3} & \best{79.4}
& \trainable{65.2} & \trainable{59.7}
& \trainable{61.6} & \trainable{58.2} \\

& LAG
& 59.5 & 58.5
& \second{59.6} & 68.1
& 59.3 & \second{81.3}
& \best{67.0} & \best{82.3}
& \trainable{77.1} & \trainable{73.0}
& \trainable{58.8} & \trainable{76.9} \\

& HeadCT
& 81.2 & 77.6
& 70.0 & 77.0
& \second{89.5} & \second{90.2}
& \best{91.5} & \best{91.6}
& \trainable{89.6} & \trainable{86.0}
& \trainable{91.5} & \trainable{92.2} \\

& Kaputt
& 55.7 & \best{55.9}
& 50.1 & \second{52.1}
& \second{57.4} & 48.3
& \best{63.0} & 51.5
& \trainable{57.1} & \trainable{56.5}
& \trainable{57.7} & \trainable{56.9} \\

\cmidrule(lr){2-14}

& Average
& \second{70.9} & 70.3
& 65.6 & 67.9
& 67.9 & \second{77.2}
& \best{74.7} & \best{79.2}
& \trainable{74.4} & \trainable{72.8}
& \trainable{72.8} & \trainable{75.4} \\

\midrule

\multirow{6}{*}{4-shot}

& MVTec
& \best{94.9} & \best{93.0}
& \second{89.7} & 89.5
& 82.4 & 89.2
& 86.4 & \second{91.8}
& \trainable{82.7} & \trainable{88.9}
& \trainable{95.3} & \trainable{93.8} \\

& MVTec LOCO
& 63.9 & 66.8
& \second{64.2} & 54.6
& 53.6 & \second{77.3}
& \best{67.4} & \best{79.9}
& \trainable{65.4} & \trainable{53.9}
& \trainable{62.8} & \trainable{62.2} \\

& LAG
& 65.0 & 66.9
& \second{67.9} & 73.0
& 60.1 & \second{81.3}
& \best{68.8} & \best{83.2}
& \trainable{77.2} & \trainable{75.8}
& \trainable{72.4} & \trainable{80.8} \\

& HeadCT
& 84.0 & 79.5
& 79.1 & \best{92.3}
& \second{90.0} & 90.7
& \best{92.0} & \second{91.4}
& \trainable{90.9} & \trainable{92.1}
& \trainable{91.8} & \trainable{92.3} \\

& Kaputt
& 55.6 & \best{58.2}
& 50.2 & \second{52.1}
& \second{56.8} & 47.2
& \best{62.7} & 51.8
& \trainable{57.4} & \trainable{57.7}
& \trainable{57.3} & \trainable{57.1} \\

\cmidrule(lr){2-14}

& Average
& \second{72.7} & 72.9
& 70.2 & 72.3
& 68.6 & \second{77.1}
& \best{75.5} & \best{79.6}
& \trainable{74.7} & \trainable{73.7}
& \trainable{75.9} & \trainable{77.2} \\

\bottomrule
\end{tabular}
}
\label{few-shot}
\vspace{-12pt}
\end{table*}

\noindent\textbf{Few-shot Anomaly Detection.}
Table~\ref{few-shot} reports the 1-, 2-, and 4-shot AD results. Among training-free baselines, AnomalyAgent achieves the best performance in the majority of dataset--shot settings and consistently obtains the highest average AUROC and $F_{1}$-$\max$ across all shot numbers, surpassing the strongest training-free baseline by up to 2.8\% AUROC and 2.5\% $F_{1}$-$\max$. Similar trends to the zero-shot setting can also be observed for both VLM-based and MLLM-based methods. Compared with conventional VLM-based FSAD methods, AnomalyAgent shows particularly strong improvements on datasets requiring contextual and relational understanding, while generic MLLM agents still exhibit clear limitations due to the lack of AD-specific reasoning guidance and calibrated normality priors. 
More importantly, AnomalyAgent exhibits stable cross-shot improvement, with performance consistently increasing from 1-shot to 4-shot. In contrast, Mem0, which equips the zero-shot version of AnomalyAgent with a generic memory module, does not reliably benefit from additional shots. This suggests that the advantage of AnomalyAgent comes not merely from memory augmentation, but from an AD-specific memory mechanism that transforms few-shot normal references into effective in-context normality priors for reasoning calibration. Although trainable methods such as InCTRL still achieve slightly stronger performance through AD-specific optimization, AnomalyAgent remains highly competitive without any training or parameter adaptation, further validating the effectiveness of memory-calibrated agentic reasoning for FSAD.
\subsection{Analysis of AnomalyAgent}
\noindent\textbf{Ablation Study.}
In this section, we analyze the key design choices in AnomalyAgent under both zero-shot and few-shot settings. Unless otherwise specified, GPT-4.1-mini is adopted as the default MLLM throughout this analysis for efficiency considerations. 
\begin{wraptable}{r}{0.52\textwidth}
\vspace{-8pt}
\centering
\caption{Ablation study on MVTec and MVTec LOCO under zero-shot and few-shot settings. The best and second-best results are highlighted in \textbf{bold} and \underline{underline}, respectively.}
\label{tab:ablation}
\setlength{\tabcolsep}{5pt}
\renewcommand{\arraystretch}{1.12}
\scalebox{0.68}{
\begin{tabular}{llcccc}
\toprule
\multirow{2}{*}{\textbf{Setting}} 
& \multirow{2}{*}{\textbf{Model}} 
& \multicolumn{2}{c}{\textbf{MVTec}} 
& \multicolumn{2}{c}{\textbf{MVTec LOCO}} \\
\cmidrule(lr){3-4} \cmidrule(lr){5-6}
& & AUROC & F1-max & AUROC & F1-max \\
\midrule
\multirow{4}{*}{Zero-shot}
& \textit{Base}                & 65.8 & 81.7 & 53.2 & 49.1 \\
& \textit{w. Ref.}              & 67.2 & 84.3 & 52.8 & 50.3 \\
& \textit{w. Ref. \& V-Tools}   & 73.4 & 86.6 & 51.7 & 48.6 \\
& \textbf{AnomalyAgent}         & \second{81.4} & \second{89.4} & \second{60.8} & \second{75.7} \\
\midrule
\multirow{3}{*}{Few-shot}
& \textit{w. Hard-Normal}       & 79.3 & 88.1 & 58.4 & 72.9 \\
& \textit{w. Calibration}       & 82.3 & 89.6 & 61.3 & 75.5 \\
& \textbf{AnomalyAgent}         & \textbf{83.5} & \textbf{90.4} & \textbf{62.0} & \textbf{77.1} \\
\bottomrule
\end{tabular}
}
\vspace{-12pt}
\end{wraptable}
% \begin{table}[t]
% \centering
% \caption{Ablation study on MVTec and MVTec LOCO under zero-shot and few-shot settings. The best and second-best results are highlighted in \textbf{bold} and \underline{underline}, respectively.}
% \label{tab:ablation}
% \setlength{\tabcolsep}{6pt}
% \renewcommand{\arraystretch}{1.15}
% \scalebox{0.6}{
% \begin{tabular}{llcccc}
% \toprule
% \multirow{2}{*}{Setting} 
% & \multirow{2}{*}{Model} 
% & \multicolumn{2}{c}{MVTec} 
% & \multicolumn{2}{c}{MVTec LOCO} \\
% \cmidrule(lr){3-4} \cmidrule(lr){5-6}
% & & AUROC & F1-max & AUROC & F1-max \\
% \midrule
% \multirow{4}{*}{Zero-shot}
% & \textit{Base}                & 65.8 & 81.7 & 53.2 & 49.1 \\
% & \textit{w. Ref.}              & 67.2 & 84.3 & 52.8 & 50.3 \\
% & \textit{w. Ref. \& V-Tools}   & 73.4 & 86.6 & 51.7 & 48.6 \\
% & \textbf{AnomalyAgent}    & \second{81.4} & \second{89.4} & \second{60.8} & \second{75.7} \\
% \midrule
% \multirow{3}{*}{Few-shot}
% & \textit{w. Hard-Normal}       & 79.3 & 88.1 & 58.4 & 72.9 \\
% & \textit{w. Calibration}       & 82.3 & 89.6 & 61.3 & 75.5 \\
% & \textbf{AnomalyAgent}    & \textbf{83.5} & \textbf{90.4} & \textbf{62.0} & \textbf{77.1} \\
% \bottomrule
% \end{tabular}
% }
% \end{table}
Table~\ref{tab:ablation} presents the ablation results on MVTec and MVTec LOCO. Under the zero-shot setting, the `\textit{Base}' model, which only contains the Planner and Reasoner, achieves limited performance, especially on the challenging dataset MVTec LOCO, where structural consistency and category-level priors are more critical. Introducing reflective reasoning (`\textit{w. Ref.}') only leads to marginal improvements, suggesting that reasoning loops alone are insufficient without additional AD-related evidence or external guidance. Further incorporating general visual tools (\ie, `\textit{w. Ref. \& V-Tools}') significantly improves the performance on MVTec, demonstrating the importance of enhanced visual evidence for the detection of simple anomalies. However, the performance on MVTec LOCO slightly drops, indicating that general visual processing alone is insufficient for the detection of complex anomalies, \eg, logical anomalies in MVTec LOCO. 
In contrast, AnomalyAgent (the zero-shot version) consistently achieves the best performance across both datasets under the zero-shot setting, suggesting that the proposed anomaly-centric toolset provides valuable anomaly priors for effective detection of both simple and complex anomalies. This also indicate that such toolset should be explicitly integrated into the reasoning workflow rather than treated as optional tools.
\setlength{\intextsep}{2pt}

\begin{wrapfigure}[17]{l}{0.51\columnwidth}
\vspace{-10pt}
\centering
\includegraphics[
width=0.50\columnwidth
]{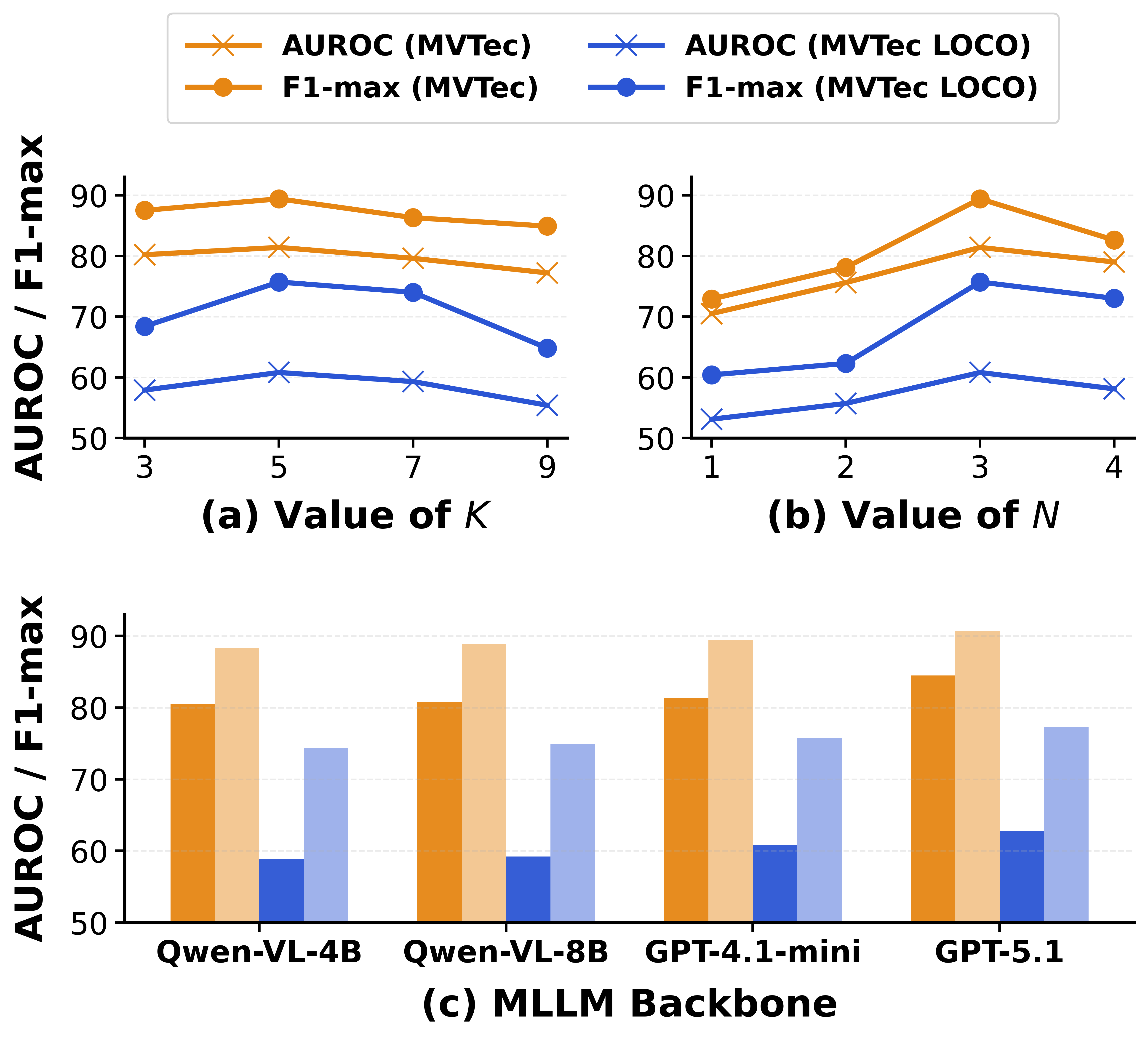}
\vspace{-10pt}
\caption{\small Hyperparameter analysis.}
\label{fig:hyp}
\vspace{-12pt}
\end{wrapfigure}
For the few-shot setting, all variants are built upon the zero-shot version of AnomalyAgent. The variant `\textit{w. Hard-Normal}' injects hard-normal information extracted from the few-shot normal references, while `\textit{w. Calibration}' uses few-shot samples to calibrate the cross-modal prior tools toward the target-data normal patterns. It is clear that using hard-normal information alone leads to only limited improvements (`\textit{w. Hard-Normal}' vs. the zero-shot AnomalyAgent), suggesting that simply providing additional normal descriptions is insufficient for effective few-shot anomaly reasoning. In contrast, the calibration variant achieves consistently stronger performance, indicating that few-shot samples are more effective when used to adapt anomaly reasoning toward the target-data normal distribution. Finally, integrating both hard-normal information and calibration into AnomalyAgent (\ie, the few-shot AnomalyAgent) yields the best overall performance, demonstrating that explicit normality priors and memory-calibrated reasoning are complementary for few-shot anomaly detection.

\noindent\textbf{Hyperparameter Sensitivity.}
% AnomalyAgent contains two key hyperparameters: $K$, which controls the number of selected atomic candidates in counterfactual template analysis, and $N$, which denotes the number of consecutive ``Normal'' judgments required for agent termination. Fig.~\ref{fig:hyp} presents the corresponding hyperparameter sensitivity analysis on MVTec and MVTec LOCO. As shown in Fig.~\ref{fig:hyp}(a), the performance first improves and then slightly drops as $K$ increases. Small $K$ values provide insufficient anomaly evidence, while overly large $K$ introduces redundant candidates that weaken counterfactual analysis. Overall, the best trade-off is achieved around $K=5$. Fig.~\ref{fig:hyp}(b) shows that increasing $N$ from 1 to 3 consistently improves the performance, indicating that multiple consecutive normal judgments help suppress premature predictions. However, excessively large $N$ slightly degrades the performance due to over-conservative reasoning. Finally, Fig.~\ref{fig:hyp}(c) shows that stronger MLLM backbones consistently lead to better anomaly detection performance, demonstrating that AnomalyAgent can effectively benefit from stronger multimodal reasoning capability.
AnomalyAgent contains two key hyperparameters: $K$, which controls the number of selected atomic candidates in counterfactual template analysis, and $N$, which denotes the number of consecutive `Normal' judgments required for agent termination.
Fig.~\ref{fig:hyp} presents the corresponding hyperparameter sensitivity analysis on MVTec and MVTec LOCO. As shown in Fig.~\ref{fig:hyp}(a), the performance first improves and then slightly drops as $K$ increases. A small $K$ may provide insufficient anomaly-relevant evidence for reliable reasoning, while an overly large $K$ can introduce redundant or noisy candidates that weaken the discriminability of the counterfactual analysis. Overall, AnomalyAgent achieves the best trade-off around $K=5$. Fig.~\ref{fig:hyp}(b) analyzes the effect of $N$. Increasing $N$ from 1 to 3 consistently improves the performance, suggesting that requiring multiple consecutive normal judgments helps suppress premature normal predictions and leads to more reliable anomaly decisions. However, further increasing $N$ to 4 slightly degrades the performance, possibly because excessive confirmation introduces over-conservative reasoning behavior. Finally, Fig.~\ref{fig:hyp}(c) compares different MLLM backbones. Larger and stronger MLLMs consistently lead to improved anomaly detection performance, verifying that the proposed reasoning framework can effectively benefit from stronger multimodal reasoning capability.

% \begin{figure}[!t]
% \centering
% \includegraphics[trim=0cm 0cm 0cm 0cm, clip, width=0.95\columnwidth]{figs/bbb.png}
% \caption{Hyperparameter Sensitivity.}
% \label{fig:framework}
% \end{figure}

\section{Conclusion}

This work introduces AnomalyAgent, offering a principled agentic framework that unleashes the superior reasoning abilities of MLLMs for training-free zero- and few-shot AD. This is not viable to the generic agentic models due to the lack of anomaly-centric priors and normality. To tackle this challenge, two novel components, including a comprehensive anomaly-centric toolset and self calibration-based memory mechanism, are introduced in AnomalyAgent. Our extensive experiments show that, with the support of these two components, AnomalyAgent achieves promising zero- and few-shot AD performance, significantly outperforming training-free VLM-based AD models and generic MLLM-based agents, with competitive performance to SotA trainable VLM-based methods.

% \clearpage

{
    \small
    \bibliographystyle{plain}
    \bibliography{references}

@article{cao2024survey,
  title={A survey on visual anomaly detection: Challenge, approach, and prospect},
  author={Cao, Yunkang and Xu, Xiaohao and Zhang, Jiangning and Cheng, Yuqi and Huang, Xiaonan and Pang, Guansong and Shen, Weiming},
  journal={arXiv preprint arXiv:2401.16402},
  year={2024}
}

@article{wu2026deep,
  title={Deep learning for video anomaly detection: A review},
  author={Wu, Peng and Pan, Chengyu and Yan, Yuting and Pang, Guansong and Yan, Qingsen and Wang, Peng and Zhang, Yanning},
  journal={IEEE Transactions on Neural Networks and Learning Systems},
  year={2026},
  publisher={IEEE}
}

@article{zheng2021generative,
  title={Generative and contrastive self-supervised learning for graph anomaly detection},
  author={Zheng, Yu and Jin, Ming and Liu, Yixin and Chi, Lianhua and Phan, Khoa T and Chen, Yi-Ping Phoebe},
  journal={IEEE Transactions on Knowledge and Data Engineering},
  volume={35},
  number={12},
  pages={12220--12233},
  year={2021},
  publisher={IEEE}
}

@inproceedings{jeong2023winclip,
  title={Winclip: Zero-/few-shot anomaly classification and segmentation},
  author={Jeong, Jongheon and Zou, Yang and Kim, Taewan and Zhang, Dongqing and Ravichandran, Avinash and Dabeer, Onkar},
  booktitle={Proceedings of the IEEE/CVF Conference on Computer Vision and Pattern Recognition},
  pages={19606--19616},
  year={2023}
}

@article{chen2023april,
  title={April-gan: A zero-/few-shot anomaly classification and segmentation method for cvpr 2023 vand workshop challenge tracks 1\&2: 1st place on zero-shot ad and 4th place on few-shot ad},
  author={Chen, Xuhai and Han, Yue and Zhang, Jiangning},
  journal={arXiv preprint arXiv:2305.17382},
  year={2023}
}

@inproceedings{zhou2024anomalyclip,
  title={AnomalyCLIP: Object-agnostic Prompt Learning for Zero-shot Anomaly Detection},
  author={Zhou, Qihang and Pang, Guansong and Tian, Yu and He, Shibo and Chen, Jiming},
  booktitle={Proceedings of The International Conference on Learning Representations},
  year={2024}
}

@inproceedings{cao2024adaclip,
  title={Adaclip: Adapting clip with hybrid learnable prompts for zero-shot anomaly detection},
  author={Cao, Yunkang and Zhang, Jiangning and Frittoli, Luca and Cheng, Yuqi and Shen, Weiming and Boracchi, Giacomo},
  booktitle={European Conference on Computer Vision},
  pages={55--72},
  year={2024}
}

@inproceedings{ma2025aa,
  title={Aa-clip: Enhancing zero-shot anomaly detection via anomaly-aware clip},
  author={Ma, Wenxin and Zhang, Xu and Yao, Qingsong and Tang, Fenghe and Wu, Chenxu and Li, Yingtai and Yan, Rui and Jiang, Zihang and Zhou, S Kevin},
  booktitle={Proceedings of the Computer Vision and Pattern Recognition Conference},
  pages={4744--4754},
  year={2025}
}

@inproceedings{fang2025af,
  title={AF-CLIP: Zero-Shot Anomaly Detection via Anomaly-Focused CLIP Adaptation},
  author={Fang, Qingqing and Lv, Wenxi and Su, Qinliang},
  booktitle={Proceedings of the 33rd ACM International Conference on Multimedia},
  pages={4846--4855},
  year={2025}
}

@inproceedings{gu2024filo,
  title={Filo: Zero-shot anomaly detection by fine-grained description and high-quality localization},
  author={Gu, Zhaopeng and Zhu, Bingke and Zhu, Guibo and Chen, Yingying and Li, Hao and Tang, Ming and Wang, Jinqiao},
  booktitle={Proceedings of the 32nd ACM International Conference on Multimedia},
  pages={2041--2049},
  year={2024}
}

@inproceedings{zhu2025fine,
  title={Fine-grained abnormality prompt learning for zero-shot anomaly detection},
  author={Zhu, Jiawen and Ong, Yew-Soon and Shen, Chunhua and Pang, Guansong},
  booktitle={Proceedings of the IEEE/CVF International Conference on Computer Vision},
  pages={22241--22251},
  year={2025}
}

@inproceedings{radford2021CLIP,
  title={Learning transferable visual models from natural language supervision},
  author={Radford, Alec and Kim, Jong Wook and Hallacy, Chris and Ramesh, Aditya and Goh, Gabriel and Agarwal, Sandhini and Sastry, Girish and Askell, Amanda and Mishkin, Pamela and Clark, Jack and others},
  booktitle={International conference on machine learning},
  pages={8748--8763},
  year={2021}
}

@inproceedings{bergmann2019mvtec,
  title={MVTec AD--A comprehensive real-world dataset for unsupervised anomaly detection},
  author={Bergmann, Paul and Fauser, Michael and Sattlegger, David and Steger, Carsten},
  booktitle={Proceedings of the IEEE/CVF conference on computer vision and pattern recognition},
  pages={9592--9600},
  year={2019}
}

@inproceedings{salehi2021multiresolution,
  title={Multiresolution knowledge distillation for anomaly detection},
  author={Salehi, Mohammadreza and Sadjadi, Niousha and Baselizadeh, Soroosh and Rohban, Mohammad H and Rabiee, Hamid R},
  booktitle={Proceedings of the IEEE/CVF Conference on Computer Vision and Pattern Recognition},
  pages={14902--14912},
  year={2021}
}

@inproceedings{li2019attention,
  title={Attention based glaucoma detection: A large-scale database and CNN model},
  author={Li, Liu and Xu, Mai and Wang, Xiaofei and Jiang, Lai and Liu, Hanruo},
  booktitle={Proceedings of the IEEE/CVF Conference on Computer Vision and Pattern Recognition},
  pages={10571--10580},
  year={2019}
}

@article{pang2021Anomaly,
author = {Pang, Guansong and Shen, Chunhua and Cao, Longbing and Hengel, Anton Van Den},
title = {Deep Learning for Anomaly Detection: A Review},
year = {2021},
volume = {54},
number = {2},
journal = {ACM Computing Surveys},
pages = {1--38},
}

@inproceedings{hofer2025kaputt,
  title={Kaputt: A large-scale dataset for visual defect detection},
  author={H{\"o}fer, Sebastian and Henning, Dorian F and Amiranashvili, Artemij and Morrison, Douglas and Tzes, Mariliza and Posner, Ingmar and Matvienko, Marc and Rennola, Alessandro and Milan, Anton},
  booktitle={Proceedings of the IEEE/CVF International Conference on Computer Vision},
  pages={24224--24233},
  year={2025}
}

@article{bergmann2022mvtecloco,
  title={Beyond dents and scratches: Logical constraints in unsupervised anomaly detection and localization},
  author={Bergmann, Paul and Batzner, Kilian and Fauser, Michael and Sattlegger, David and Steger, Carsten},
  journal={International Journal of Computer Vision},
  volume={130},
  number={4},
  pages={947--969},
  year={2022},
  publisher={Springer}
}

@inproceedings{zhu2024inctrl,
  title={Toward generalist anomaly detection via in-context residual learning with few-shot sample prompts},
  author={Zhu, Jiawen and Pang, Guansong},
  booktitle={Proceedings of the IEEE/CVF conference on computer vision and pattern recognition},
  pages={17826--17836},
  year={2024}
}

@inproceedings{xu2026mrad,
  title={MRAD: Zero-Shot Anomaly Detection with Memory-Driven Retrieval},
  author={Xu, Chaoran and Lv, Chengkan and Chen, Qiyu and Zhang, Feng and Zhang, Zhengtao},
  booktitle={Proceedings of The International Conference on Learning Representations},
  year={2026}
}

@inproceedings{yao2023react,
  title={REACT: SYNERGIZING REASONING AND ACTING IN LANGUAGE MODELS},
  author={Yao, Shunyu and Zhao, Jeffrey and Yu, Dian and Du, Nan and Shafran, Izhak and Narasimhan, Karthik and Cao, Yuan},
  booktitle={Proceedings of The International Conference on Learning Representations},
  year={2023}
}

@article{shinn2023reflexion,
  title={Reflexion: Language agents with verbal reinforcement learning},
  author={Shinn, Noah and Cassano, Federico and Gopinath, Ashwin and Narasimhan, Karthik and Yao, Shunyu},
  journal={Advances in neural information processing systems},
  volume={36},
  pages={8634--8652},
  year={2023}
}

@inproceedings{zhang2025logsad,
  title={Towards training-free anomaly detection with vision and language foundation models},
  author={Zhang, Jinjin and Wang, Guodong and Jin, Yizhou and Huang, Di},
  booktitle={Proceedings of the IEEE/CVF Conference on Computer Vision and Pattern Recognition},
  pages={15204--15213},
  year={2025}
}

@inproceedings{jiang2025mmad,
  title={MMAD: A Comprehensive Benchmark for Multimodal Large Language Models in Industrial Anomaly Detection},
  author={Jiang, Xi and Li, Jian and Deng, Hanqiu and Liu, Yong and Gao, Bin-Bin and Zhou, Yifeng and Li, Jialin and Wang, Chengjie and Zheng, Feng},
  booktitle={Proceedings of The International Conference on Learning Representations},
  year={2025}
}

@inproceedings{xu2025anomalyOV,
  title={Towards zero-shot anomaly detection and reasoning with multimodal large language models},
  author={Xu, Jiacong and Lo, Shao-Yuan and Safaei, Bardia and Patel, Vishal M and Dwivedi, Isht},
  booktitle={Proceedings of the Computer Vision and Pattern Recognition Conference},
  pages={20370--20382},
  year={2025}
}

@article{chao2025anomalyr1,
  title={Anomalyr1: A grpo-based end-to-end mllm for industrial anomaly detection},
  author={Chao, Yuhao and Liu, Jie and Tang, Jie and Wu, Gangshan},
  journal={arXiv preprint arXiv:2504.11914},
  year={2025}
}

@article{chhikara2025mem0,
  title={Mem0: Building production-ready ai agents with scalable long-term memory},
  author={Chhikara, Prateek and Khant, Dev and Aryan, Saket and Singh, Taranjeet and Yadav, Deshraj},
  journal={arXiv preprint arXiv:2504.19413},
  year={2025}
}

@inproceedings{qu2025dictas,
  title={Dictas: A framework for class-generalizable few-shot anomaly segmentation via dictionary lookup},
  author={Qu, Zhen and Tao, Xian and Gong, Xinyi and Qu, ShiChen and Zhang, Xiaopei and Wang, Xingang and Shen, Fei and Zhang, Zhengtao and Prasad, Mukesh and Ding, Guiguang},
  booktitle={Proceedings of the IEEE/CVF International Conference on Computer Vision},
  pages={20519--20528},
  year={2025}
}

@inproceedings{roth2022patchcore,
  title={Towards total recall in industrial anomaly detection},
  author={Roth, Karsten and Pemula, Latha and Zepeda, Joaquin and Sch{\"o}lkopf, Bernhard and Brox, Thomas and Gehler, Peter},
  booktitle={Proceedings of the IEEE/CVF conference on computer vision and pattern recognition},
  pages={14318--14328},
  year={2022}
}

@article{bai2023qwen,
  title={Qwen technical report},
  author={Bai, Jinze and Bai, Shuai and Chu, Yunfei and Cui, Zeyu and Dang, Kai and Deng, Xiaodong and Fan, Yang and Ge, Wenbin and Han, Yu and Huang, Fei and others},
  journal={arXiv preprint arXiv:2309.16609},
  year={2023}
}

@article{li2024llava,
  title={Llava-onevision: Easy visual task transfer},
  author={Li, Bo and Zhang, Yuanhan and Guo, Dong and Zhang, Renrui and Li, Feng and Zhang, Hao and Zhang, Kaichen and Zhang, Peiyuan and Li, Yanwei and Liu, Ziwei and others},
  journal={arXiv preprint arXiv:2408.03326},
  year={2024}
}

@article{liu2024deepseek,
  title={Deepseek-v3 technical report},
  author={Liu, Aixin and Feng, Bei and Xue, Bing and Wang, Bingxuan and Wu, Bochao and Lu, Chengda and Zhao, Chenggang and Deng, Chengqi and Zhang, Chenyu and Ruan, Chong and others},
  journal={arXiv preprint arXiv:2412.19437},
  year={2024}
}

@article{ResAD,
      title={ResAD: A Simple Framework for Class Generalizable Anomaly Detection}, 
      author={Xincheng Yao and Zixin Chen and Gao Chao and Guangtao Zhai and Chongyang Zhang},
      year={2024},
      booktitle={Thirty-Eighth Annual Conference on Neural Information Processing Systems, NeurIPS 2024},
      url={https://arxiv.org/abs/2410.20047},
      primaryClass={cs.CV}
}

@article{li2024promptad,
  title={PromptAD: Learning Prompts with only Normal Samples for Few-Shot Anomaly Detection},
  author={Li, Xiaofan and Zhang, Zhizhong and Tan, Xin and Chen, Chengwei and Qu, Yanyun and Xie, Yuan and Ma, Lizhuang},
  journal={Proceedings of the IEEE/CVF conference on computer vision and pattern recognition},
  year={2024}
}

@article{gu2023anomalyagpt,
  title={AnomalyGPT: Detecting Industrial Anomalies using Large Vision-Language Models},
  author={Gu, Zhaopeng and Zhu, Bingke and Zhu, Guibo and Chen, Yingying and Tang, Ming and Wang, Jinqiao},
  journal={Proceedings of the AAAI Conference on Artificial Intelligence},
  year={2024}
}

@article{fuvcka2026anomalyvfm,
  title={AnomalyVFM--Transforming Vision Foundation Models into Zero-Shot Anomaly Detectors},
  author={Fu{\v{c}}ka, Matic and Zavrtanik, Vitjan and Sko{\v{c}}aj, Danijel},
  journal={arXiv preprint arXiv:2601.20524},
  year={2026}
}

@misc{openai2025gpt51,
  title        = {GPT-5.1},
  author       = {{OpenAI}},
  year         = {2025},
  howpublished = {\url{https://platform.openai.com/docs/models}},
}

@misc{openai2025gpt41,
  title        = {GPT-4.1 Mini},
  author       = {{OpenAI}},
  year         = {2025},
  howpublished = {\url{https://platform.openai.com/docs/models}},
}

@article{yang2025qwen3,
  title   = {Qwen3 Technical Report},
  author  = {Yang, An and others},
  journal = {arXiv preprint arXiv:2505.xxxxx},
  year    = {2025}
}

@inproceedings{wang2021real,
  title={Real-esrgan: Training real-world blind super-resolution with pure synthetic data},
  author={Wang, Xintao and Xie, Liangbin and Dong, Chao and Shan, Ying},
  booktitle={Proceedings of the IEEE/CVF international conference on computer vision},
  pages={1905--1914},
  year={2021}
}
}

\clearpage

\appendix

\section{More Implmentation Details}

\subsection{Details of Visual Tool Use}
\label{app:tool_details}

AnomalyAgent equips the MLLM agent with a set of lightweight visual tools to support anomaly-oriented evidence gathering. These tools provide complementary observations under different visual conditions, and their outputs are not used as standalone anomaly predictions. Instead, the processed observations are returned to the planner--reasoner--reflector loop, where they are considered together with the original image, semantic priors, counterfactual evidence, and accumulated reasoning history. All tools are invoked only at inference time and do not require target-dataset training or parameter adaptation.

\paragraph{Image Denoising.}
The denoising tool is implemented with OpenCV's fast non-local means filtering, which suppresses color and spatial noise while preserving local image structures. In AnomalyAgent, this tool is useful when suspected defects may be confused with acquisition noise, compression artifacts, or low-light degradation, allowing the reasoner to verify whether the abnormal evidence remains visible after noise suppression.

\paragraph{Image Deblurring.}
The deblurring tool is implemented with unsharp masking based on Gaussian blur subtraction. It enhances local edges and contours without introducing any learned adaptation. This tool is used when anomaly hypotheses depend on boundary sharpness, object shape, or surface discontinuities, helping the agent distinguish true structural irregularities from blur-induced ambiguity.

\paragraph{Brightness Enhancement.}
The brightness enhancement tool applies contrast-limited adaptive histogram equalization (CLAHE) to the L-channel in the LAB color space. This operation improves local contrast while limiting over-amplification. In the reasoning loop, it is used to re-examine regions affected by under-exposure or over-exposure, where abnormal patterns may otherwise be hidden or visually exaggerated.

\paragraph{Image Super-Resolution.}
The super-resolution tool uses Real-ESRGAN~\cite{wang2021real} to enhance image resolution and recover fine-grained texture details. This provides additional visual evidence for cases where the anomaly decision relies on small defects, subtle texture changes, or local missing components. The enhanced image is treated as an auxiliary observation for verification rather than as a replacement for the original image.

\paragraph{Image Retrieval.}
The retrieval tool uses CLIP-based visual-textual retrieval~\cite{radford2021CLIP}, where cosine similarity between CLIP embeddings is used to score relevance. This tool provides contextual reference evidence by retrieving previously observed examples related to the current query or suspected abnormal pattern. The retrieved results are used to support comparison with normal or suspicious visual patterns, but the final anomaly decision is still made by the reasoner through the agentic reasoning process.

\section{Prompt Templates}
\label{app:prompts}

\begin{figure}[!t]
\centering
\includegraphics[trim=0cm 0cm 0cm 0cm, clip, width=\columnwidth]{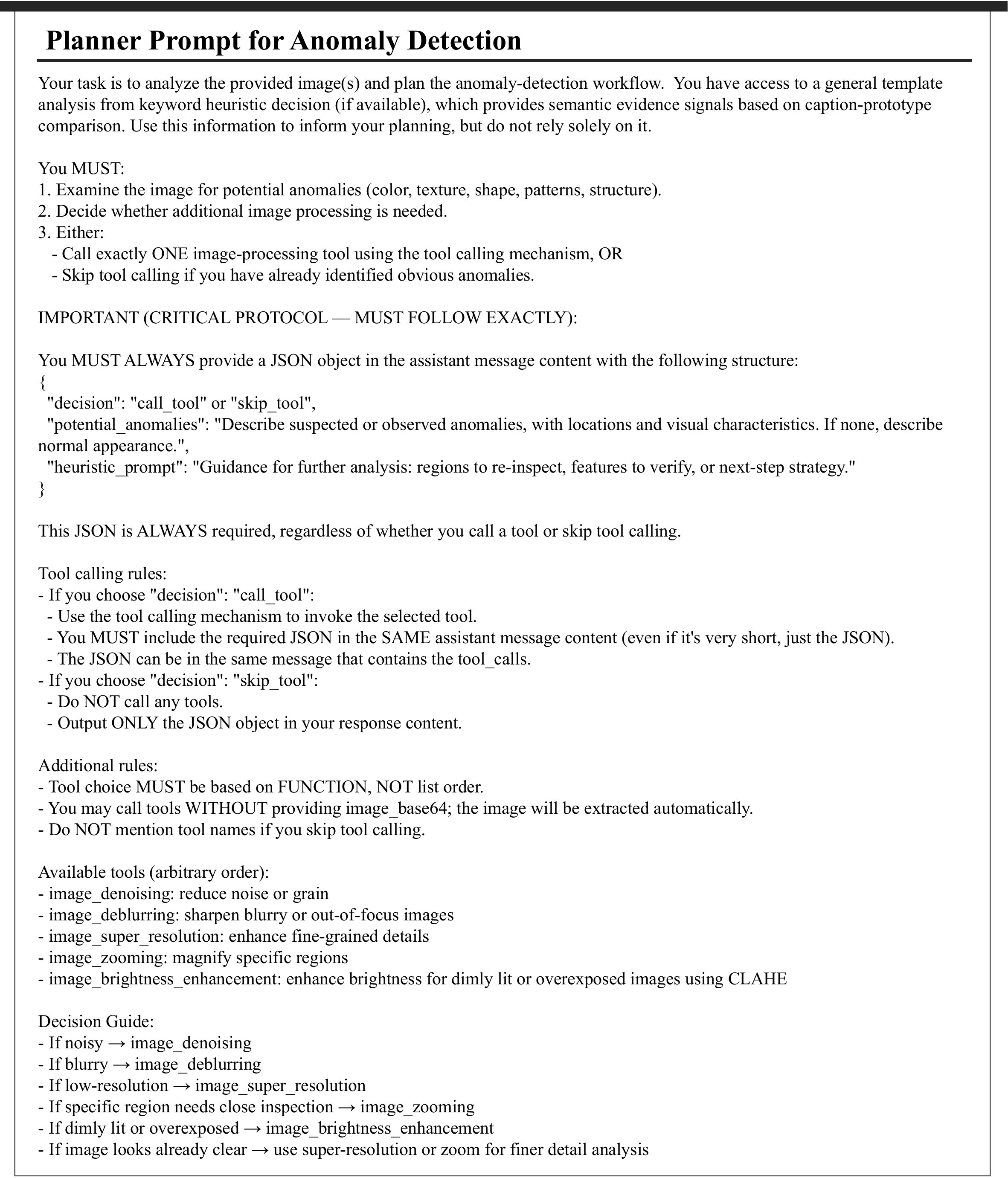}
\caption{Planner Prompt for Anomaly Detection.}
\end{figure}

\begin{figure}[!t]
\centering
\includegraphics[trim=0cm 0cm 0cm 0cm, clip, width=\columnwidth]{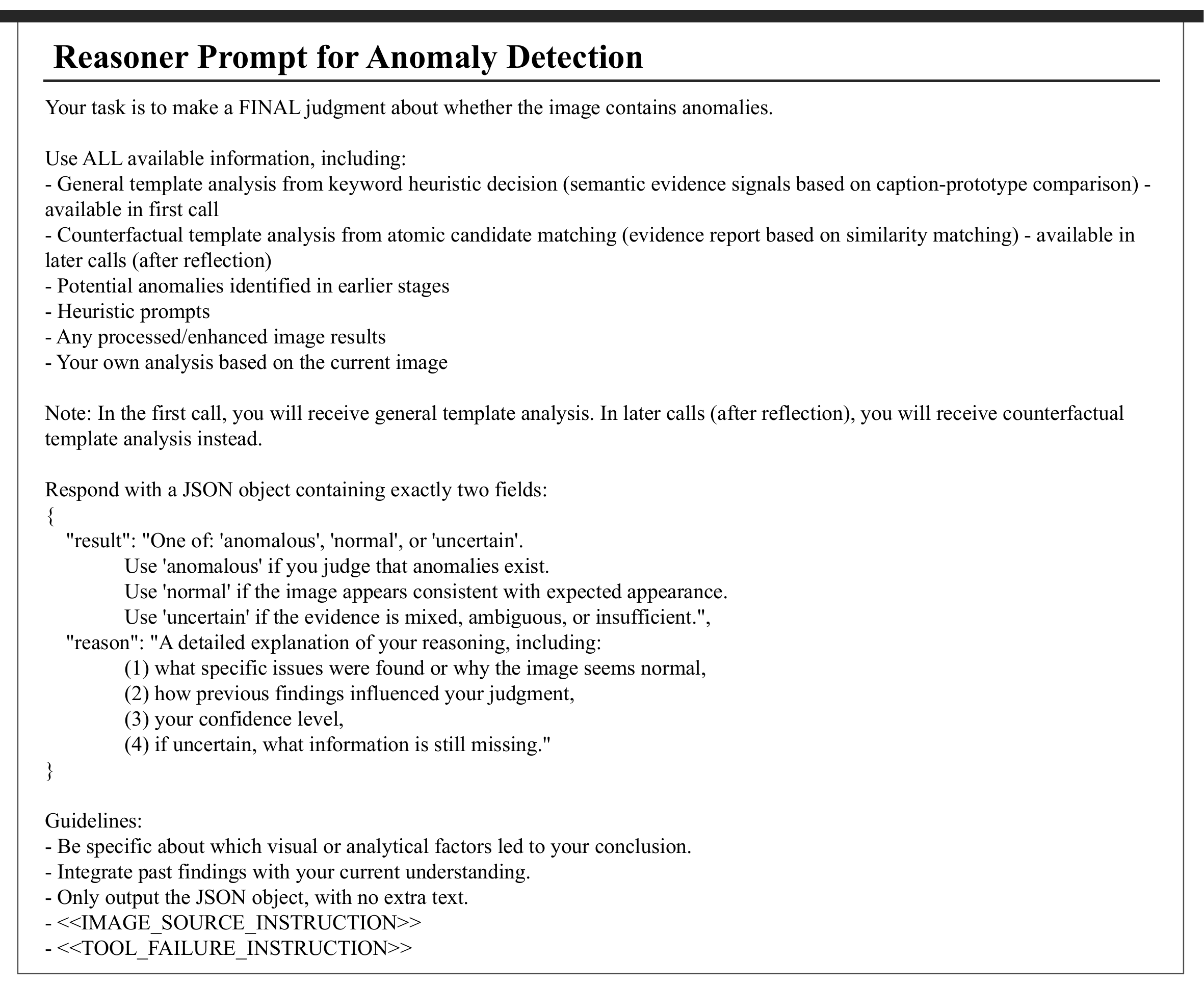}
\caption{Reasoner Prompt for Anomaly Detection.}
\end{figure}

\begin{figure}[!t]
\centering
\includegraphics[trim=0cm 0cm 0cm 0cm, clip, width=\columnwidth]{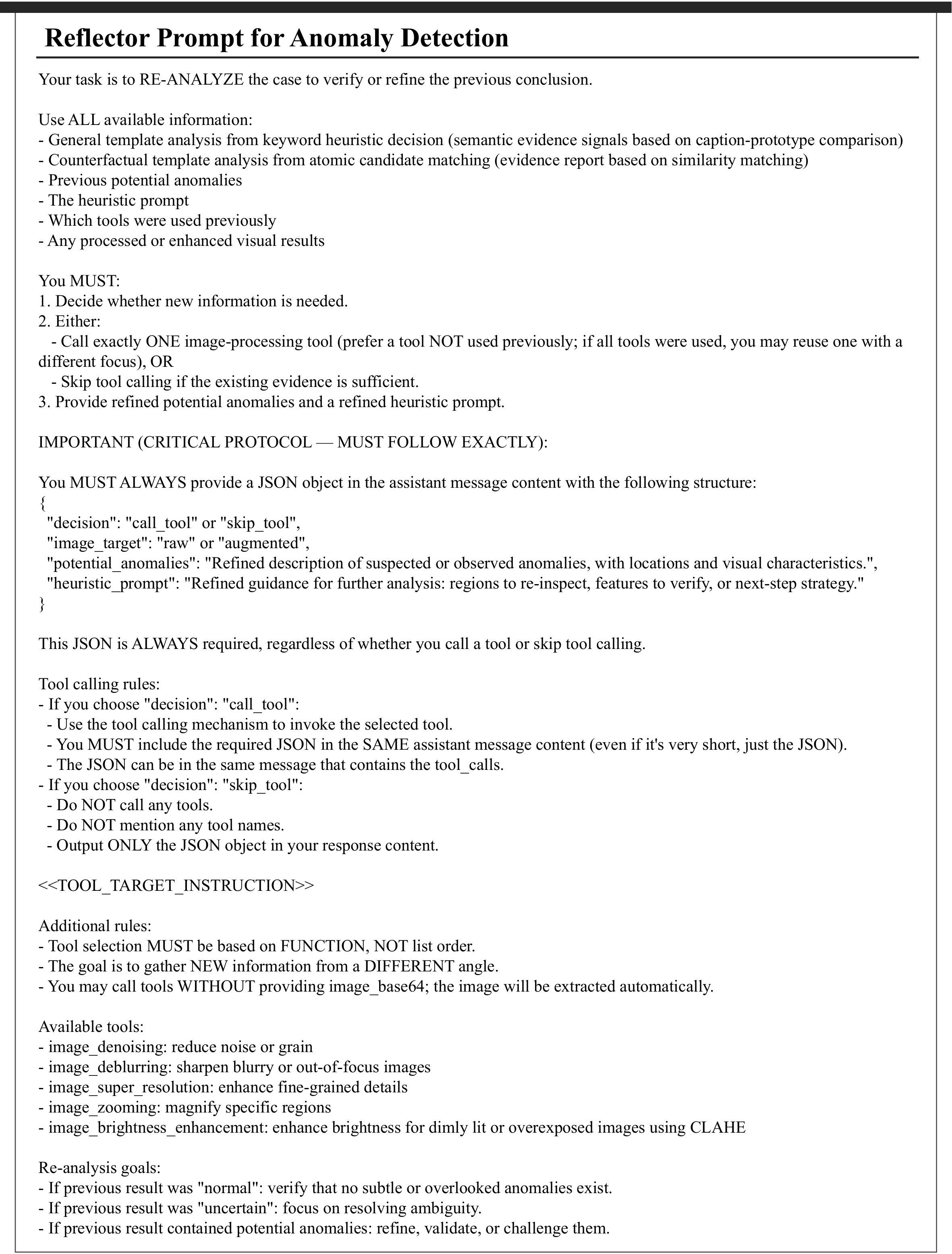}
\caption{Reflector Prompt for Anomaly Detection.}
\end{figure}

\begin{figure}[!t]
\centering
\includegraphics[trim=0cm 0cm 0cm 0cm, clip, width=\columnwidth]{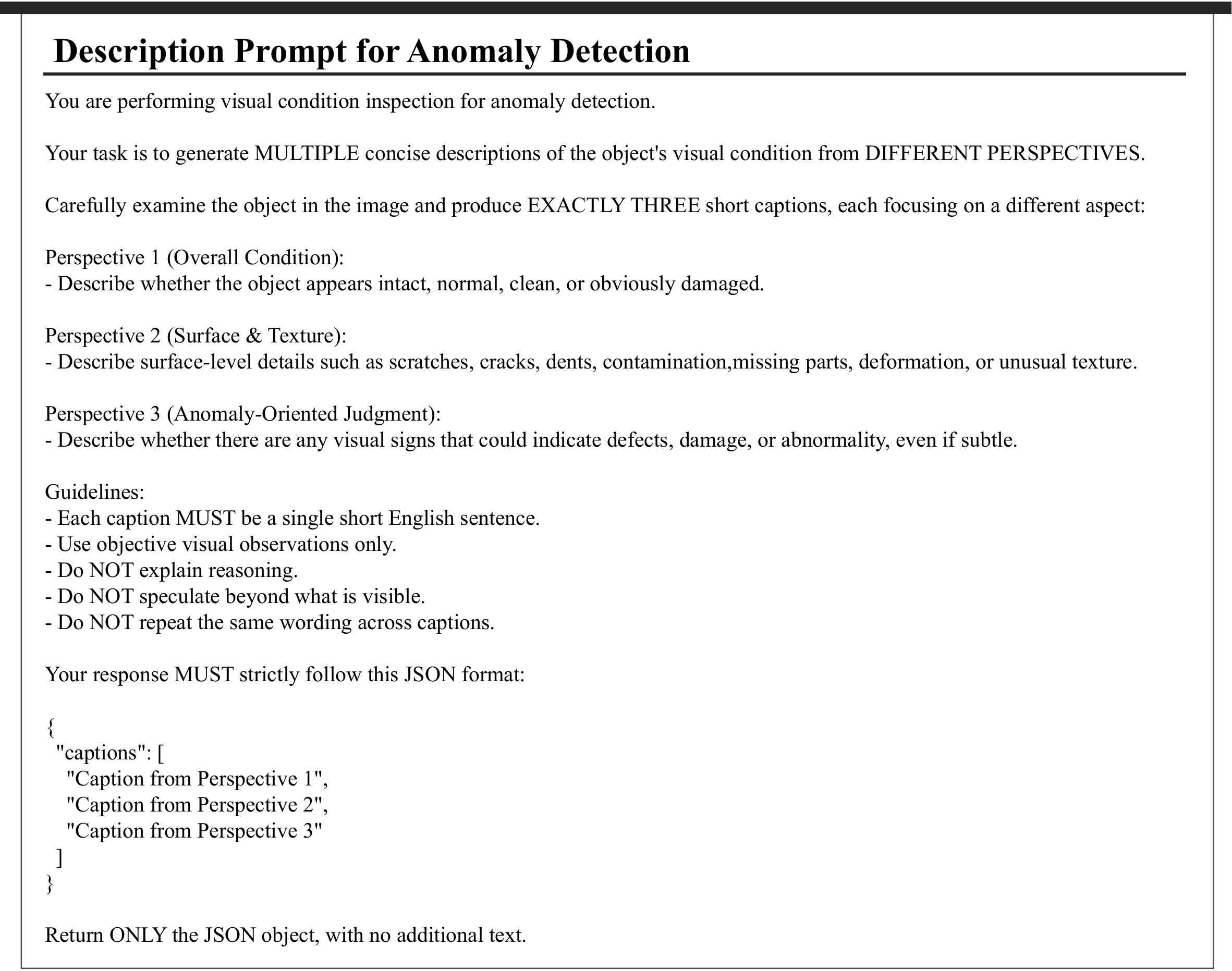}
\caption{Description Prompt for Anomaly Detection.}
\end{figure}

\begin{figure}[!t]
\centering
\includegraphics[trim=0cm 0cm 0cm 0cm, clip, width=\columnwidth]{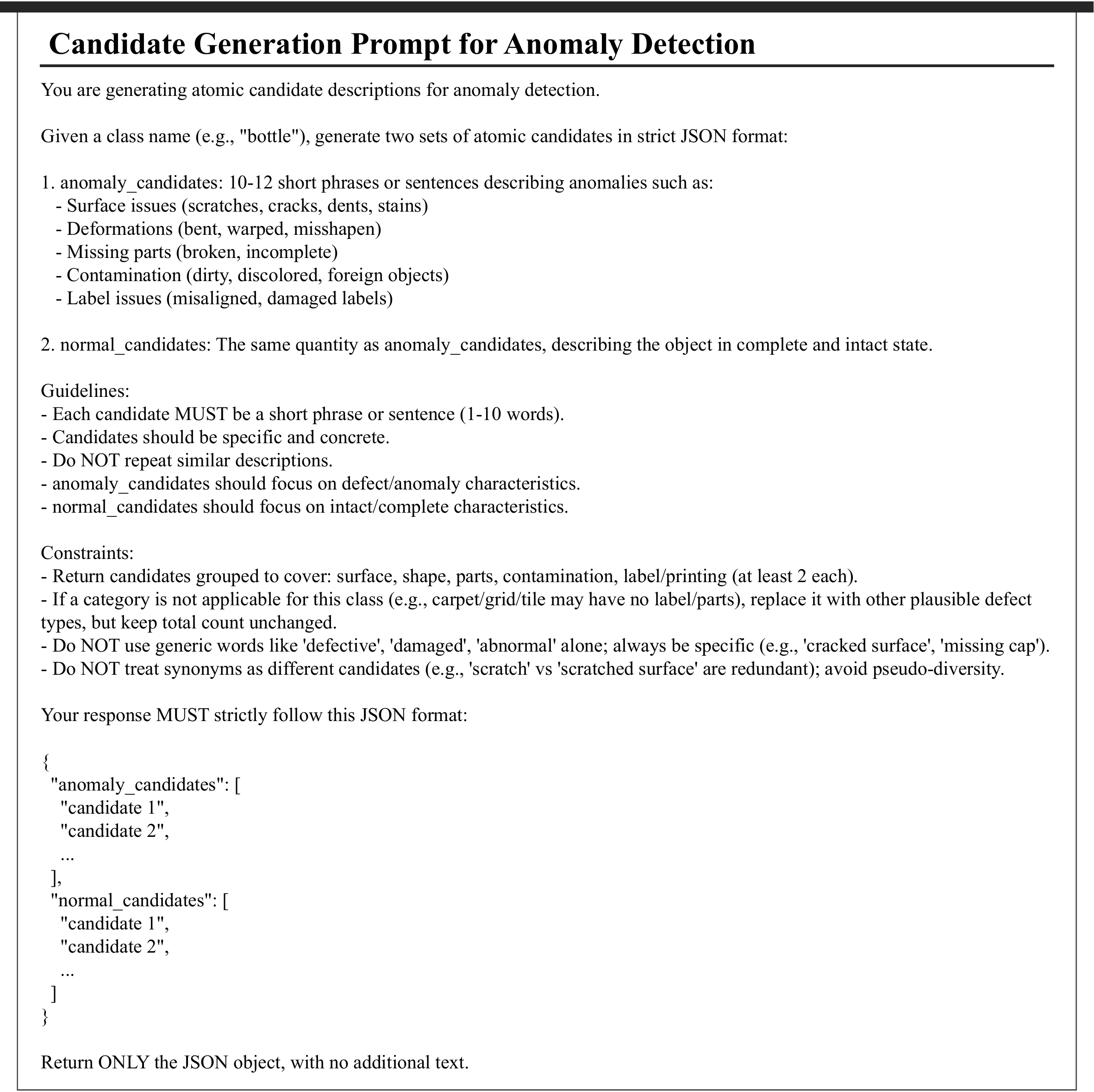}
\caption{Candidate Generation Prompt for Anomaly Detection.}
\end{figure}

\begin{figure}[!t]
\centering
\includegraphics[trim=0cm 0cm 0cm 0cm, clip, width=\columnwidth]{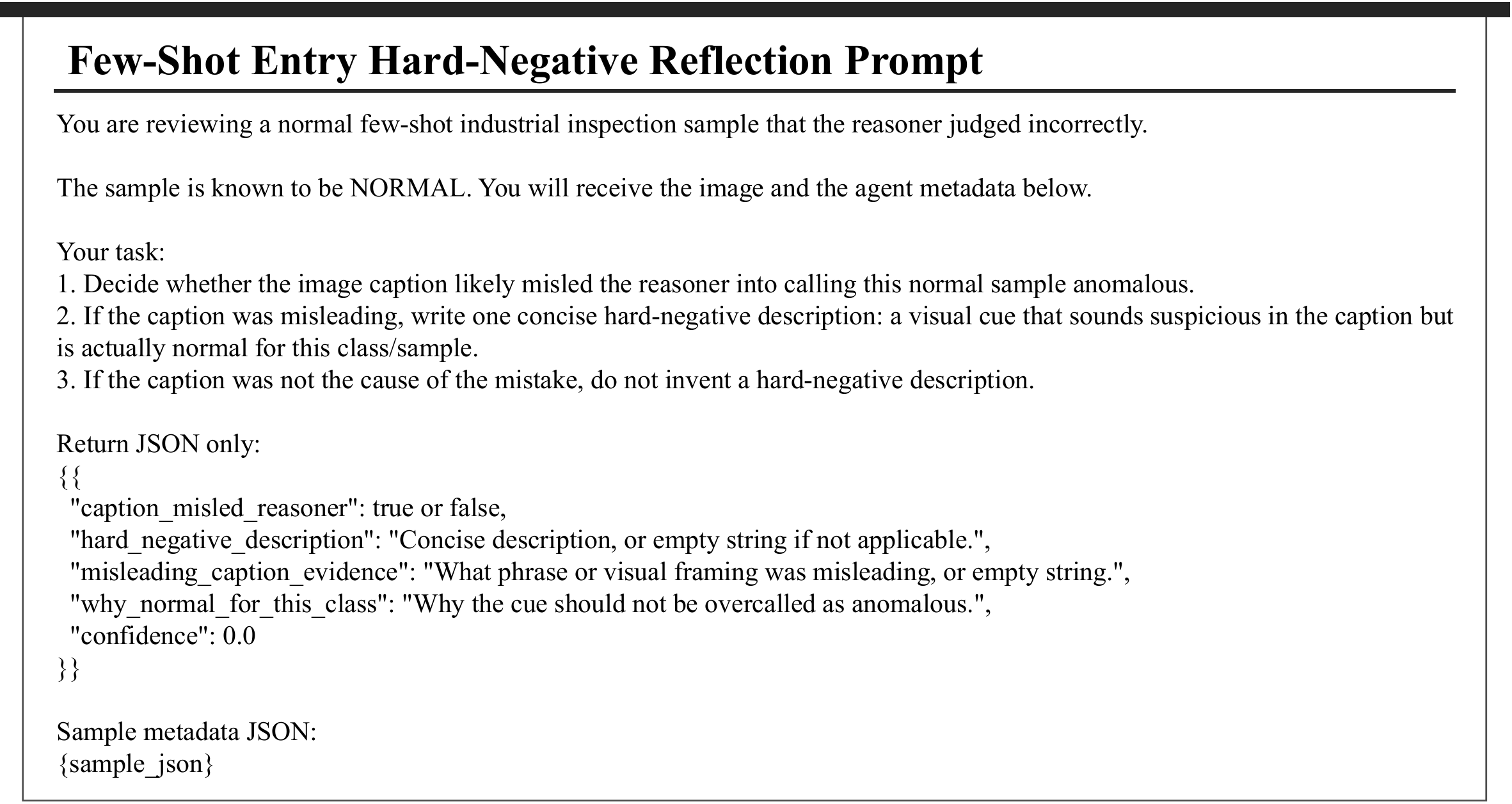}
\caption{Few-Shot Entry Hard-Negative Reflection Prompt.}
\end{figure}

\begin{figure}[!t]
\centering
\includegraphics[trim=0cm 0cm 0cm 0cm, clip, width=\columnwidth]{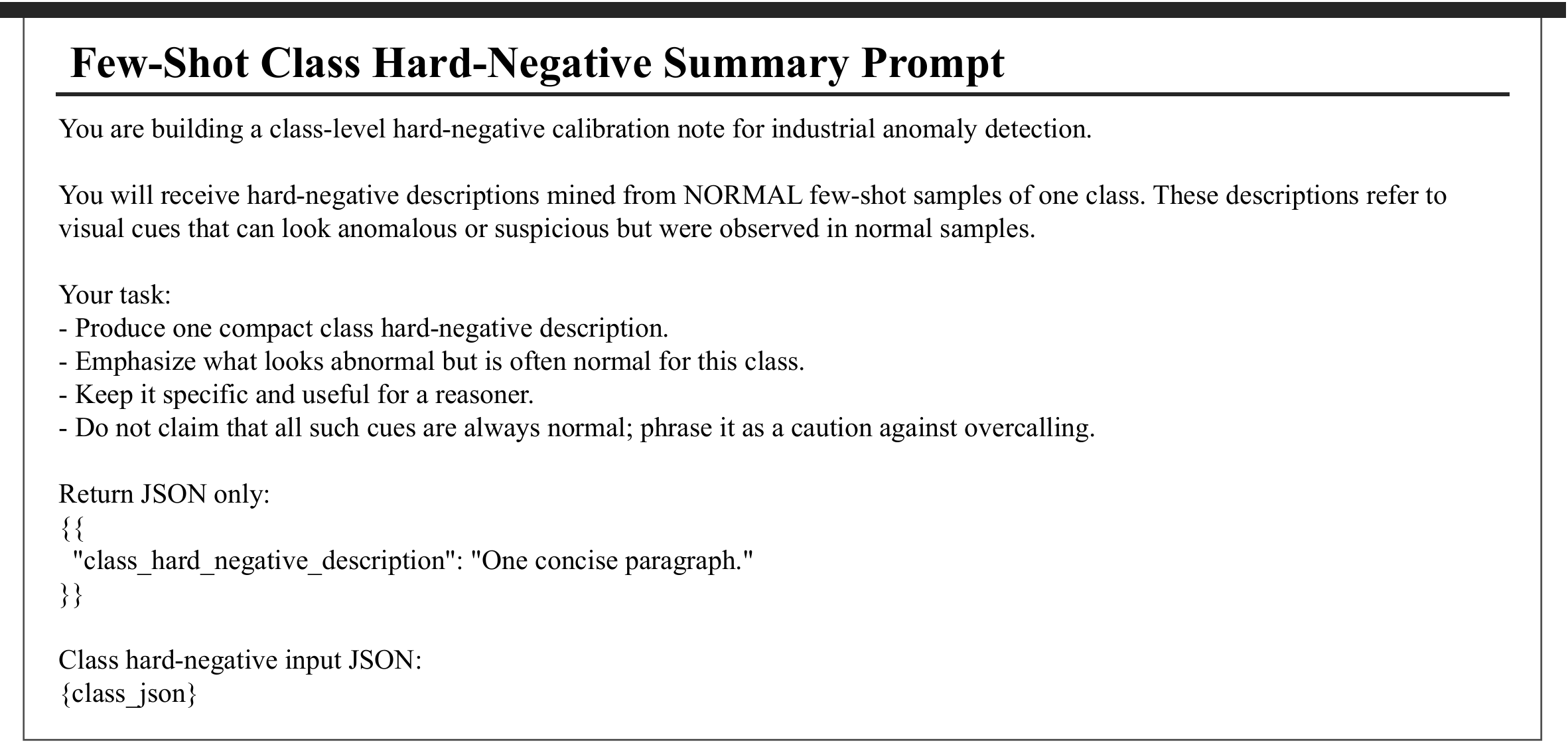}
\caption{Few-Shot Class Hard-Negative Summary Prompt.}
\end{figure}

\clearpage

\end{document}